\documentclass[journal]{IEEEtran}
\usepackage[numbers,sort&compress]{natbib}
\usepackage{booktabs}

\usepackage{url}
\usepackage{algorithmic}
\usepackage{algorithm}
\usepackage{multirow}
\usepackage{amssymb}
\usepackage[table ]{ xcolor}
\usepackage{threeparttable}
\usepackage{float}
\usepackage[misc]{ifsym}

\ifCLASSINFOpdf
\usepackage{amsmath}
\usepackage[pdftex]{graphicx}
\graphicspath{{../pdf/}{../jpeg/}}
\DeclareGraphicsExtensions{.pdf,.jpeg,.png}
\else
\usepackage[dvips]{graphicx}
\usepackage{pifont}       
\usepackage{bbding}       
\usepackage{fontawesome}
\graphicspath{{../eps/}}
\fi
\begin{document}
	
	\title{SFFNet: A Wavelet-Based Spatial and Frequency Domain Fusion Network for Remote Sensing Segmentation}
	
	\author{Yunsong~Yang, Genji~Yuan, and~Jinjiang~Li
		\thanks{Y. Yang, G. Yuan and J. Li are with School of Computer Science and
			Technology, Shandong Technology and Business University, Yantai 264005,
			China}
	}

	\maketitle
	
	\begin{abstract}
		  In order to fully utilize spatial information for segmentation and address the challenge of handling areas with significant grayscale variations in remote sensing segmentation, we propose the SFFNet (Spatial and Frequency Domain Fusion Network) framework. This framework employs a two-stage network design: the first stage extracts features using spatial methods to obtain features with sufficient spatial details and semantic information; the second stage maps these features in both spatial and frequency domains. In the frequency domain mapping, we introduce the Wavelet Transform Feature Decomposer (WTFD) structure, which decomposes features into low-frequency and high-frequency components using the Haar wavelet transform and integrates them with spatial features. To bridge the semantic gap between frequency and spatial features, and facilitate significant feature selection to promote the combination of features from different representation domains, we design the Multiscale Dual-Representation Alignment Filter (MDAF). This structure utilizes multiscale convolutions and dual-cross attentions. Comprehensive experimental results demonstrate that, compared to existing methods, SFFNet achieves superior performance in terms of mIoU, reaching 84.80\% and 87.73\% respectively.The code is located at \url{https://github.com/yysdck/SFFNet}.
		  
	\end{abstract}
	
	
	\IEEEpeerreviewmaketitle
	
	\section{Introduction}
	
	\IEEEPARstart{W}ith the continuous development of sensors and aerospace technology, high-resolution satellite and aerospace remote sensing images can be easily obtained, providing high-resolution observations of diverse landscapes on Earth, covering various scenes from urban areas to farmlands, forests, and lakes. Remote sensing image segmentation is a crucial technology aimed at dividing remote sensing images of Earth into different objects or categories, which is essential for geographic information systems (GIS), resource management, environmental monitoring, and crisis management. Here are some key applications utilizing remote sensing image segmentation,such as land cover mapping \cite{mapping1,mapping2}, change detection \cite{detection1,detection2}, environmental protection \cite{protection1,protection2}, road and building extraction \cite{extraction1,extraction2}, and many other practical applications \cite{applicitions1,applications2}.
	
	\begin{figure}
		\includegraphics[width=0.48\textwidth]{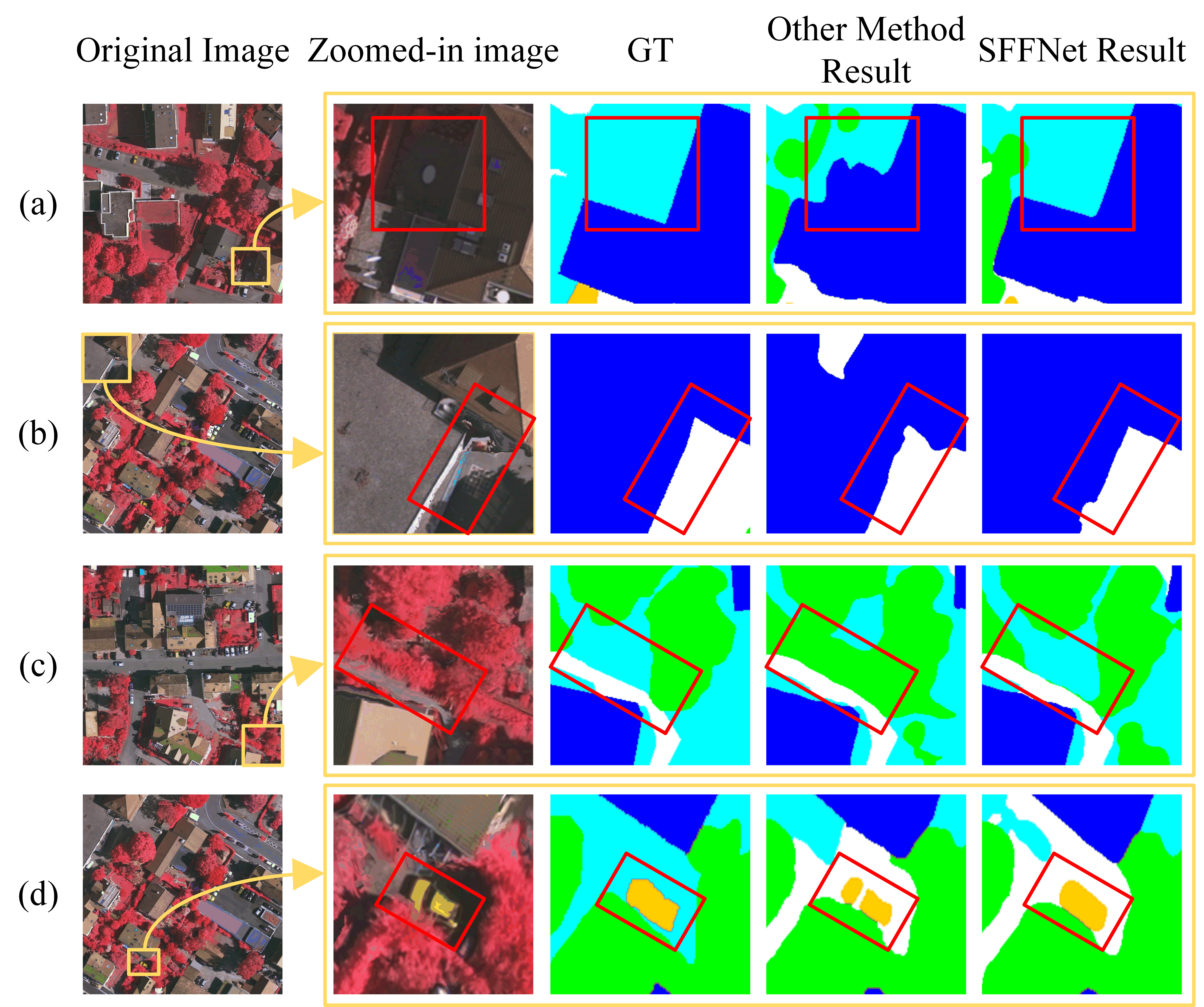}
		\caption{The figure illustrates the challenges in remote sensing image segmentation: areas with large grayscale variations (such as shadows, edges, and regions with significant texture changes) are difficult to accurately segment, and the sole use of frequency domain features leads to spatial information loss. The first column shows the original images, the second column shows locally enlarged images, the third column shows the locally enlarged ground truth labels (GT), the fourth column displays segmentation results of some mainstream methods, and the fifth column presents the segmentation results of SFFNet. Specifically: (a) showcases the segmentation results of ST-Unet and SFFNet in shadow areas. (b) shows the segmentation results of ST-Unet and SFFNet in edge areas. (c) demonstrates the segmentation results of ST-Unet and SFFNet (our method) in regions with significant texture changes. (d) displays a scenario where XNet segments a car into two halves due to spatial information loss, while SFFNet offers improvement. From (a) to (c), it can be observed that segmentation methods not utilizing frequency domain features perform poorly in handling areas with large grayscale variations, while (d) illustrates the issue of spatial information loss caused by solely frequency domain-based methods.}
		\label{fig:1}       
	\end{figure}
	
	In the field of remote sensing segmentation, various deep learning-based methods have been proposed, covering multiple data types, including hyperspectral images (HSI), optical images, LiDAR data, and integrated multisensor data \cite{md1,md2}. Convolutional neural network (CNN) methods have been proven effective in classifying and segmenting each pixel in a given image into semantic labels \cite{lables}. There are still many scholars researching CNN-based segmentation networks, such as AFF-Unet \cite{affunet}, CATNet \cite{catnet}.
	
	Although CNNs perform well in overall segmentation accuracy, they have limitations in handling complex object characteristics in remote sensing images, such as small scale, high similarity, mutual occlusion, etc. \cite{transunet}. Therefore, it is crucial to globally model remote sensing segmentation networks. The current mainstream approach is to use attention mechanisms to establish long-range dependencies, such as Dual Attention \cite{dual}, Criss-cross structure \cite{ccnet}, and multiscale attention \cite{mscsanet}. However, these methods still rely on transforming global information into local aggregations of CNNs, rather than directly encoding global context, making it difficult to obtain clear global scene information from complex remote sensing images. In contrast, the self-attention mechanism of Transformers can avoid this problem. Subsequent scholars introduced Vit \cite{vit} and Swintransformer \cite{swintransformer} to adapt to computer vision tasks, which also play an important role in remote sensing image segmentation. Popular segmentation network designs in remote sensing image segmentation often combine Transformers and CNNs, such as ST-Unet \cite{stunet} and CSTUnet \cite{cstunet}
	
	In the field of remote sensing image segmentation, these CNN and Transformer fusion networks can effectively utilize global and local information to some extent\cite{xu2019lecture2note,xu2022semantic}. However, these methods are based solely on spatial segmentation and do not utilize frequency domain information. In the complex background of remote sensing images, grayscale variations pose challenges to spatial segmentation. Particularly, regions with significant grayscale variations such as edges and shadows lead to segmentation errors, while frequency domain features are more sensitive to these areas\cite{xu2022morphtext,xu2024seeing}. Therefore, introducing frequency domain information into spatial segmentation networks is necessary.
	
	Haar wavelet transform, as a fast decomposition method, is often used for frequency domain transformation in image processing \cite{imageprocessing}. It is commonly employed for tasks such as decomposition \cite{decomposition}, compression \cite{comparison}, denoising \cite{denoising}, among others. A few studies have applied it to image segmentation \cite{haarsegmentation1,haarsegmentation2,xnet}, where these methods directly substitute frequency domain features from Haar wavelet transform for spatial domain features, thus improving the model's performance and generalization to some extent. However, due to the particularity of remote sensing images, the sole use of frequency domain features leads to unclear information and spatial information loss. In contrast, spatial domain features are more capable of capturing semantic information of different categories in images and more accurate spatial information. Therefore, a method is needed to effectively introduce frequency domain information while maintaining spatial domain features to address the challenges of spatial segmentation and improve segmentation model accuracy.
	
	Based on the above ideas, we designed a Spatial and Frequency Domain Fusion Network (SFFNet), adopting a two-stage approach to preserve rich semantic information and spatial information of spatial domain features while introducing additional frequency domain features. Firstly, spatial domain feature extraction is performed in the first stage, followed by feature mapping utilizing spatial features from the first stage, including global feature mapping and local feature mapping, to retain sufficient spatial information. Simultaneously, to introduce frequency domain features, we propose a Wavelet Transform Feature Decomposer (WTFD) module, utilizing Haar wavelet transform to decompose spatial features into high-frequency and low-frequency signals, which are then converted into high-frequency and low-frequency features embedded into CNN networks to complement mapped global and local features. Subsequently, to achieve cross-scale alignment and feature selection, we design a Wavelet Transform Feature Decomposer (MDAF) module, which utilizes multiscale vertical bar convolution and dual-cross attention to implement Dual-Representation Alignment Filter (DAF) for semantic alignment and feature selection. Through these methods, we expand local and global features with frequency domain information, enhancing feature representation capability, enabling the model to more comprehensively consider areas with large grayscale variations in images, such as shadows, edges, and regions with significant texture changes, thereby improving segmentation accuracy and robustness. Additionally, to enable Swin Transformer to achieve more efficient global feature mapping to adapt to remote sensing segmentation tasks, we establish remote dependencies between Windows using a set of vertical convolutions.
		
	In summary, the contributions of this paper are as follows:
	
	1.~Proposal of Wavelet Transform Feature Decomposer(WTFD): We introduce a Haar wavelet transform decomposer as an additional frequency domain feature mapping branch, utilizing it to decompose spatial features into high-frequency and low-frequency information to supplement local and global information in the network, thereby preserving the spatial characteristics and rich semantic features of spatial domain features and effectively introducing additional frequency domain information.
	
	2.~Proposal of Multiscale Dual-Representation Alignment Filter(MDAF): To achieve cross-scale semantic alignment and feature selection between frequency domain information and spatial domain information, we design an MDAF module, which effectively bridges the semantic gap between frequency domain features and spatial domain features and selects more representative features for segmentation through multiscale vertical convolution and dual-cross attention, implementing Dual-Representation Alignment Filter (DAF).
	
	3.~Proposal of Spatial and Frequency Domain Fusion Network (SSFNet): Based on the above structures, we propose a two-stage SSFNet network architecture that preserves spatial information and rich semantic information of spatial domain features. The first stage performs spatial feature extraction, while the second stage ensures sufficient spatial information through global feature mapping and local feature mapping, while introducing additional frequency domain information through WTFD. MDAF achieves semantic alignment and feature selection between frequency domain features and spatial domain features, enabling the model to more comprehensively consider features. Experimental results demonstrate that this network architecture effectively addresses the challenges of spatial segmentation and achieves excellent performance in two mainstream remote sensing segmentation datasets.

	\begin{figure*}
		\center{}
		\includegraphics[width=1\textwidth]{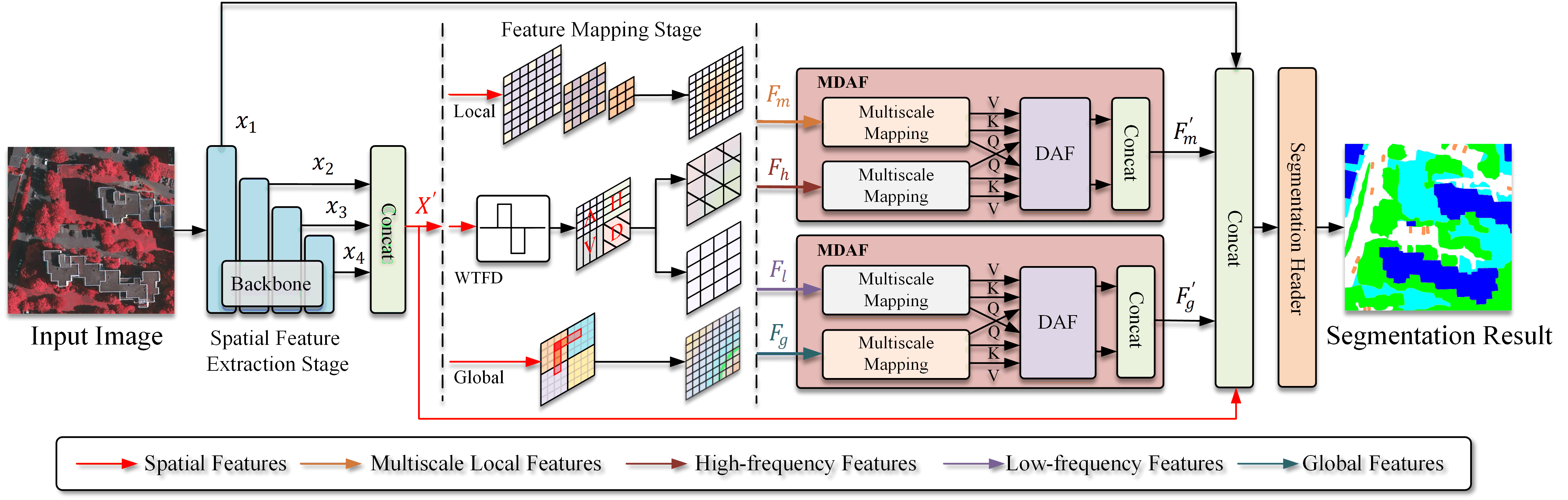}
		\caption{The main framework of SFFNet illustrates a two-stage segmentation network. The first stage involves spatial feature extraction to acquire sufficient spatial information. Subsequently, various feature mappings are performed in the second stage, including global feature mapping, local feature mapping, and frequency domain feature mapping. Specifically, global feature mapping and local feature mapping preserve diverse spatial information, while frequency domain feature mapping introduces additional frequency domain information. The frequency domain feature mapping is achieved through the WTFD method, followed by alignment of spatial and frequency domain features using MDAF, bridging their semantic gaps and facilitating the combination of both features.}
		\label{fig:2}       
	\end{figure*}

	\section{Related work}

	Remote sensing segmentation is a critical task in handling remote sensing images, aiming to partition images into different regions based on land cover classes or features. Researchers strive to enhance the accuracy and efficiency of segmentation results by exploring various methods and technologies. Successful semantic segmentation in remote sensing has significant applications in urban planning, agriculture monitoring, among others. This section will introduce some important works related to remote sensing segmentation and discuss their impact and contributions to this field.

	\subsection{CNN-based Remote Sensing Image Semantic Segmentation}
	
	Traditional approaches in remote sensing image segmentation emphasize the design of robust features compatible with spectral information and local image textures \cite{textures1,textures2}, such as the method proposed by Huang et al \cite{huang}. Despite high-resolution datasets providing clear geometric information and fine textures\cite{ft}, they also introduce more noise, increasing the segmentation difficulty.
	
	Deep learning techniques are widely applied in remote sensing segmentation. Fully Convolutional Networks (FCNs) \cite{fcn} address semantic segmentation problems, but their performance is limited \cite{segmentationresearch1,segmentationresearch2}. UNet \cite{unet} focuses on refining semantic segmentation tasks and has become a standard model. Some improvement methods such as ResUnet-a \cite{resunet} and Refine-UNet \cite{refineunet} enhance segmentation accuracy.
	
	Due to the specificity of remote sensing images, such as complex backgrounds, small targets, and occlusions, traditional methods often struggle to improve segmentation accuracy due to their limited receptive fields. To address this issue, researchers attempt to apply multiscale features. Spatial pyramid pooling in Deeplab \cite{deeplab} is a multiscale feature extraction method, and other networks like MSCANet \cite{mscanet} and MSLANet \cite{mslanet} also utilize multiscale features.
	
	In this paper, we first utilize CNN for spatial feature extraction and then employ a multi-scale downsampling CNN branch to perform local multiscale mapping of features in order to obtain multiscale features and retain spatial diversity of local feature scales.

	\subsection{Transformer-based Semantic Segmentation of Remote Sensing Images}
	
	To overcome the limitations of local patterns in CNNs, attention mechanisms play a crucial role in remote sensing image segmentation. Li et al.\cite{resunet} proposed a linear attention mechanism, while Ding et al. \cite{ding2020lanet} designed local attention blocks, and MSCSA-Net \cite{mscsanet} utilized local channel spatial attention and multiscale attention. Although these attention mechanisms improve performance, they still overly rely on convolutional operations without directly encoding global context.
	
	With the introduction of Vit \cite{vit}, the above issue is avoided, and numerous studies have emerged, such as SwinTransformer \cite{swintransformer}, providing new ideas for directly encoding global context. Currently, remote sensing image segmentation methods generally adopt Transformer-based approaches, including pure Transformer architectures such as Segmenter \cite{strudel2021segmenter} and SwinUNet \cite{swinunet}, as well as models that combine Transformer and CNN, such as GLOTS \cite{GLOTS} and EMRT \cite{EMRT}.
	
	However, directly using SwinTransformer or Vit for global feature mapping involves high computational complexity. In remote sensing tasks, a lower computational approach for global feature mapping should be adopted. In this paper, we integrate SwinTransformer as an independent branch into the CNN network to retain global spatial information. Moreover, to better adapt to remote sensing tasks, we replace SW-MSA with vertical convolutions for window interactions, achieving a more efficient global feature mapping.
	
	\begin{figure*}
		\centering
		\includegraphics[width=0.95\textwidth]{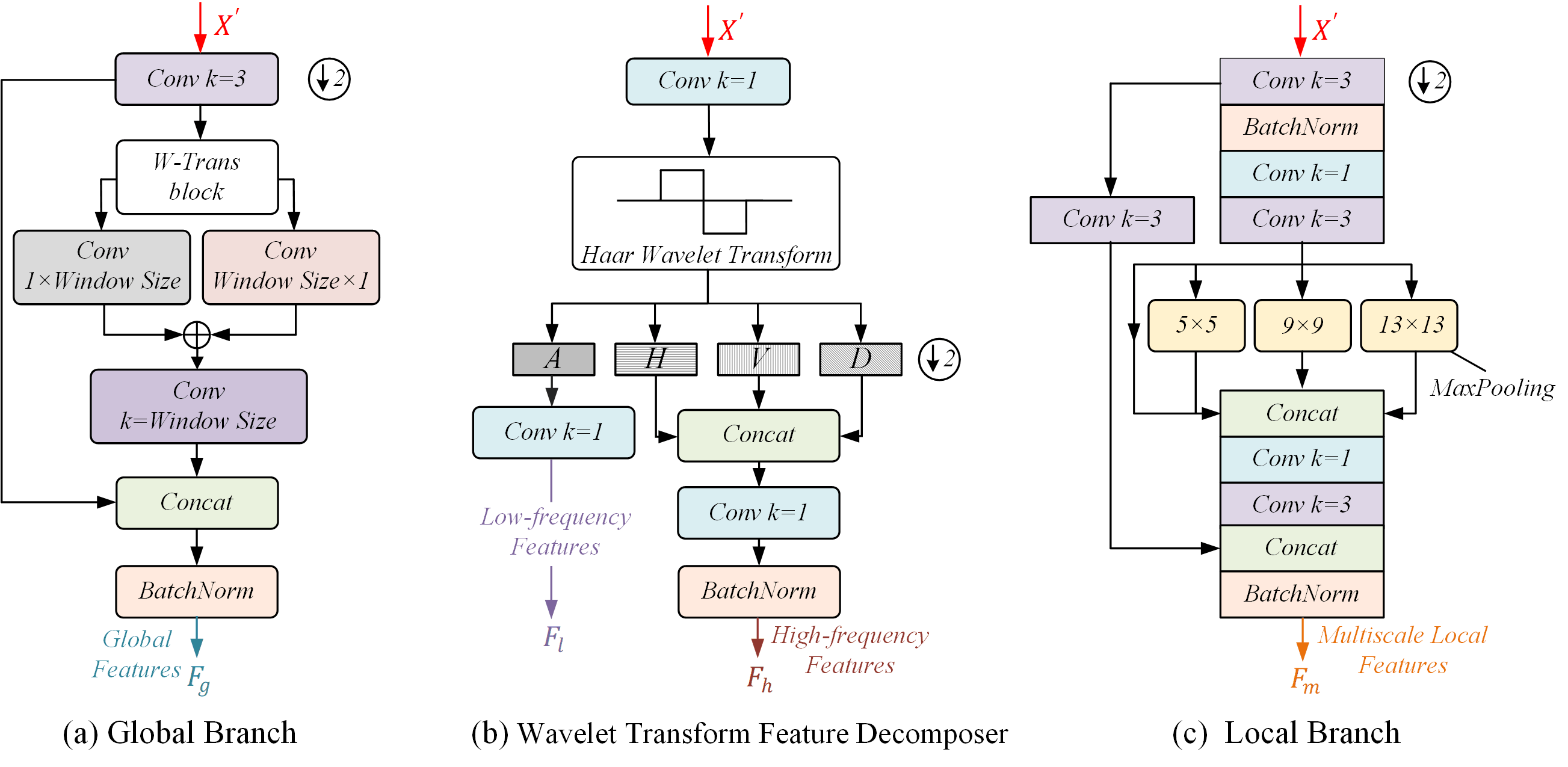}
		\caption{This Figure illustrates three feature mapping methods in the feature mapping stage. (a) represents the global branch, which is used to map the original features into global features. This module replaces the Shift window in Swintransformer with vertical bar-shaped convolutions, enabling the improved module to have efficient global modeling capabilities and better adaptability to remote sensing tasks. (b) represents WTFD (Wavelet Transform Feature Decomposer), which provides frequency domain information to the network. This structure decomposes the original features into low-frequency and high-frequency features using the Haar wavelet transform, and subsequently combines them with spatial features to enable the model to consider features in a new representation domain. (c) represents the local branch, which utilizes pooling pyramids to map the original features into spatially local multi-scale features.}
		\label{fig:3}       
	\end{figure*}

	\subsection{Wavelet Transform in Image Processing}
	
	Wavelet transforms play a significant role in signal and image analysis, providing a multi-resolution representation while capturing the low-frequency and high-frequency components of signals or images, aiding in better understanding their structure and features. In image processing, methods combining CNNs and wavelet transforms have been widely used in tasks such as image restoration, super-resolution, and compression. For example, in \cite{liu2018multi}, a multi-level wavelet CNN (MWCNN) model was proposed for image restoration. In \cite{huang2017wavelet}, a wavelet-based CNN (Wavelet-SRNet) was introduced for multiscale face super-resolution. In \cite{ma2019iwave}, a wavelet-like transform integrated into CNNs for image compression. These studies fully leverage the advantages of CNN feature learning and the multi-resolution analysis provided by wavelet technology.

	In the segmentation field, methods combining CNNs and wavelet transforms have also been studied \cite{haarsegmentation1,haarsegmentation2,xnet}. These methods typically directly utilize purely frequency domain features for segmentation. However, in remote sensing image segmentation, due to the specificity of remote sensing images, solely using frequency domain features often leads to the loss of crucial spatial information. To balance the advantages of spatial information and frequency domain features, we propose SFFNet, which adopts a two-stage approach for segmentation. We first perform spatial feature extraction, and then in the second stage, we introduce WTFD to introduce additional frequency domain features, decompose spatial features into high-frequency and low-frequency information, and embed them into the CNN network. Simultaneously, we map the original features to global and local features to retain rich spatial information. Subsequently, we use MDAF to bridge the semantic gap between frequency domain features and spatial features, achieve feature selection, and fuse the features of the two representation domains. Such a design aims to retain sufficient spatial information while addressing the challenge of handling areas with large grayscale variations in spatial image segmentation.

	\section{Method}
	
	In this section, we will first outline the overall structure of SFFNet. Next, we will discuss several key components of SFFNet in detail, namely the Global branch, Local branch, and WTFD used for frequency domain feature mapping in the feature mapping stage, as well as the MDAF used for alignment and selection of frequency domain features and spatial domain features. Finally, we will introduce the loss function adopted in our approach.

	\subsection{SFFNet Structure}
	
	The structure of SFFNet is depicted in Fig.\ref{fig:2}, designed as a two-stage segmentation network. The first stage utilizes ConvNext \cite{convnext} for sufficient spatial feature extraction, adjusting features from the last three downsampling stages to the same scale and merging them. The second stage is the feature mapping stage, transforming raw data with redundant information into a more representative and discriminative feature space through feature mapping. There are three branches for feature mapping: the Global branch utilizes an improved Swin Transformer for global feature mapping, the Local branch is used for multi-scale local feature mapping, and the Wavelet Transformer branch utilizes Haar wavelet transform to map original features into frequency-domain high-frequency and low-frequency features. Subsequently, MDAF is used to align, select, and combine local features and high-frequency features, as well as global features and low-frequency features, to obtain more powerful mixed features. Finally, the expanded global features, local features, and mixed features from the feature extraction stage are merged, and the final segmentation results are obtained through the segmentation header.
	
	Specifically, for an input image $x \in \mathbb{R}^{3 \times h \times w}$, where $h$ and $w$ are the height and width of the input image, respectively, the first stage of feature extraction yields four features of different scales: $x_1 \in \mathbb{R}^{C \times H \times W}$, $x_2 \in \mathbb{R}^{2C \times (H/2) \times (W/2)}$, $x_3 \in \mathbb{R}^{4C \times (H/4) \times (W/4)}$, $x_4 \in \mathbb{R}^{8C \times (H/8) \times (W/8)}$, where $H$ and $W$ are the downsampled height and width, and $C = 96$. Then, through convolution and interpolation, the latter three features are merged, expressed as follows:
	
	\begin{equation}
	\label{eq:1}
	X' = Cat(\delta_{1 \times 1}(\varphi(x_2)), \delta_{1 \times 1}(\varphi(x_3)), \delta_{1 \times 1}(\varphi(x_4)))
	\end{equation}
	
	where $\varphi(\cdot)$ represents the interpolation operation, $\delta_{k \times k}(\cdot)$ denotes the convolution with a kernel size of $k \times k$, $Cat$ indicates the concatenation operation, and $X' \in \mathbb{R}^{3C \times (H/2) \times (W/2)}$ represents the fused features of $x_2$, $x_3$, and $x_4$. This fusion operation effectively combines deep semantic information and surface spatial information, providing more powerful spatial features for subsequent operations.
	
	Subsequently, $X'$ is used as the raw data for the second stage feature mapping. Through the three feature mapping branches in HWT, frequency-domain high-frequency and low-frequency features, as well as spatial global and local features, are obtained:
	
	\begin{equation}
	\label{eq:2}
	\left\{ \begin{array}{l}
		F_g = f_g(X')\\
		F_m = f_m(X')\\
		F_l, F_h = f_w(X')\\
	\end{array} \right. 
	\end{equation}
	
	where $f_g(\cdot)$, $f_m(\cdot)$, and $f_w(\cdot)$ denote the operations of the Global branch, Local branch, and WTFD for feature mapping, respectively. $F_g$, $F_m$, $F_l$, and $F_h$ represent the mapped global features, multi-scale local features, low-frequency features, and high-frequency features, respectively, with dimensions $C \times (H/4) \times (W/4)$.
	
	Next, the spatial features and frequency-domain features are split into two groups and fed into MDAF, which aligns the semantic gaps between frequency-domain features and spatial features for feature selection. Finally, these two features, along with the first-layer output of the feature extraction stage, are combined, and the final segmentation results are obtained through the segmentation header:
	
	\begin{equation}
	\label{eq:3}
	\left\{ \begin{array}{l}
	F_{g}^{'} = f_{mdaf}(F_g, F_l)\\
	F_{m}^{'} = f_{mdaf}(F_m, F_h)\\
	\end{array} \right. 
	\end{equation}
	
	\begin{equation}
	\label{eq:4}
		Y = \delta(Cat(\varphi(F_{g}^{'}), \varphi(F_{m}^{'}), x_1, \delta(X^{'})))
	\end{equation}
	
	where $f_{mdaf}(S, F)$ denotes the MDAF block for aligning spatial features $S$ and frequency-domain features $F$, $F_{g}^{'}$ and $F_{m}^{'}$ represent the global features and multi-scale local features with frequency-domain information, both with dimensions of $ (C, H/4, W/4)$, and $Y$ represents the final mixed features, with dimensions of $(C, H/2, W/2)$, which are obtained through the final segmentation header.
	
	SFFNet adopts a concise design, avoiding a complex decoder. By using a two-stage method, it retains spatial information while expanding the representation space, thereby improving the accuracy and robustness of the segmentation network. Next, we will introduce the important components of SFFNet, namely the Global branch, Local branch, WTFD, and MDAF used for achieving multiscale semantic alignment and feature selection between frequency-domain and spatial features during the feature mapping stage.
	
	\subsection{Feature Mapping}
	
	\subsubsection{Global Branch}
	\begin{figure}
		\centering
		\includegraphics[width=0.47\textwidth]{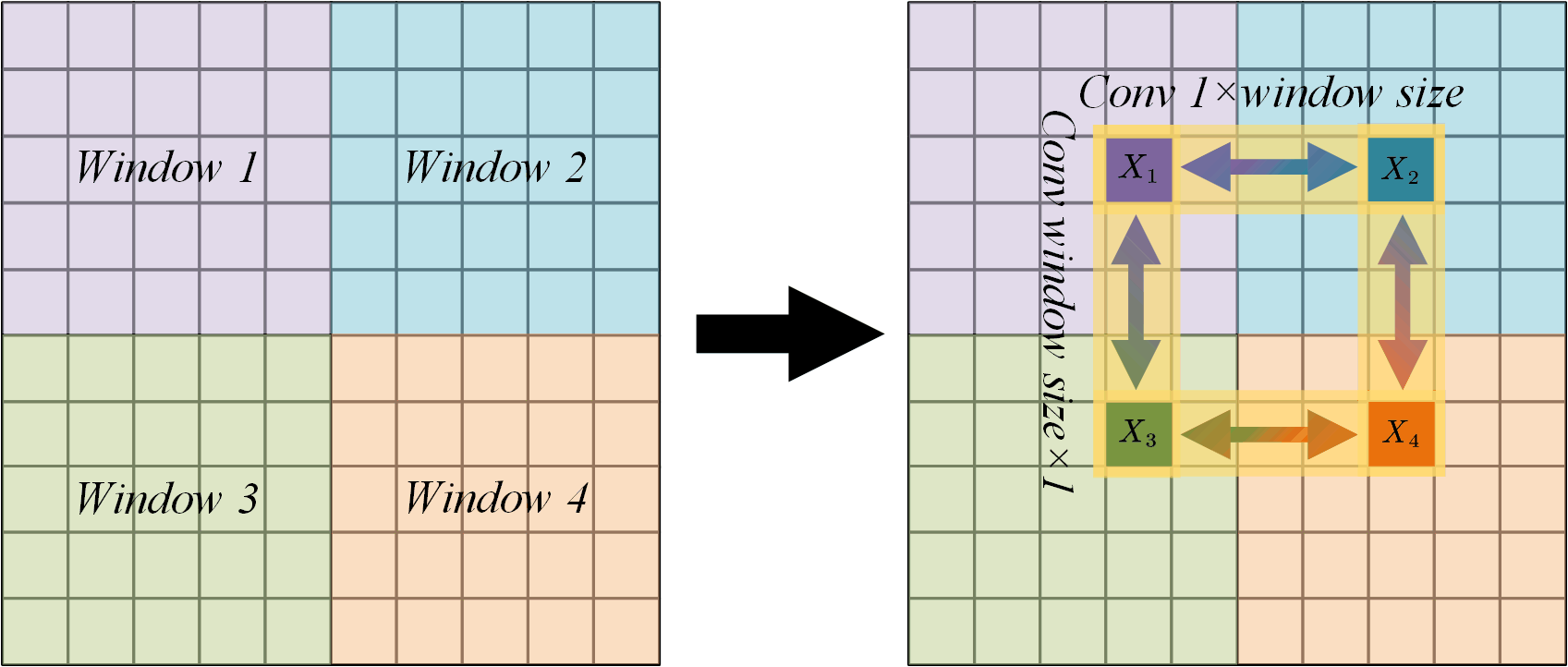}
		\caption{Establishing remote dependencies between windows using vertical stripwise convolutions. Remote dependencies between windows are established by employing a set of vertical stripwise convolutions on pre-segmented windows. For instance, as illustrated in the figure, feature connections from $x_1$ to both $x_2$ and $x_3$ are achieved through convolutions of length equal to the window size. Additionally, dependencies between $x_2$ and $x_3$ to $x_4$ are established, with each pixel possessing information from other pixels within its respective window, thereby facilitating interaction between windows.}
		\label{fig:4}       
	\end{figure}

	To make Swintransformer more efficient for global mapping and better adapt to remote sensing segmentation tasks, we have enhanced it by designing a Global branch, as illustrated in Fig.\ref{fig:3}(a). In this branch, we first downsample the features and then, similar to Swintransformer, use the W-Trans block for window partitioning and establishing global features within each window, as described in \cite{swintransformer}. However, for inter-window information interaction, we have abandoned the complex SW-Trans block in Swintransformer and instead opted for a more efficient vertical stripe convolution. The vertical stripe convolution is designed as a set of convolutions with kernel sizes of $(1\times x)$ and $(x\times1)$. Fig.\ref{fig:4} illustrates the principle of using vertical stripe convolution to establish interaction between windows. For instance, considering a feature pixel $x_1$ in window 1 that has been partitioned using the previous W-Trans block, by applying vertical stripe convolution with kernel sizes of $(1\times ws)$ and $(ws\times 1)$, where ws represents window size, we can establish remote dependencies between $x_1$ and feature pixels $x_2$ and $x_3$. Since $x_2$ and $x_3$ have already established remote dependencies with $x_4$ in window 4 using the same method, $x_1$ thus establishes remote dependencies with all other window pixels in this manner.
	
	Specifically, for an input feature $x \in \mathbb{R}^{C\times H\times W}$, we first downsample it using a 3$\times$3 convolution and then process it through the W-Trans block:
	
	\begin{equation}
	\label{eq:5}
	F_p=F_{wtrans}(\delta_{3\times3}(x))
	\end{equation}
	
	where $F_{wtrans}(\cdot)$ denotes the operation of using the W-Trans block on the features, detailed derivation can be seen in \cite{swintransformer}, and $F_p \in \mathbb{R}^{C\times(H/2)\times(W/2)}$ represents the features obtained through window partitioning and window self-attention operations.
	
	The entire global branch can be described as follows:
	\begin{equation}
	\begin{split}
	\label{eq:6}
	F_g & =f_g(x) \\
	& =\psi(Cat(\delta_{3\times3}(x),\delta_{ws\times ws}+\delta_{1\times ws}(F_p)+\delta_{ws\times1}(F_p)))
	\end{split}	
	\end{equation}

	where ws denotes the window size and $\psi(\cdot)$ denotes batch normalization. $F_g \in \mathbb{R}^{(C/3)\times(H/2)\times(W/2)}$ represents the computation result of the global branch.
	
	By using a set of vertical stripe convolutions instead of the SW-Trans block, we have achieved efficient global mapping. Moreover, objects in remote sensing images often exhibit directional features, such as roads, rivers, and edges. Using vertical stripe convolutions in the Global branch not only allows us to learn features in both horizontal and vertical directions but also enables better capture of these directional features, thus better adapting to remote sensing images.

	\subsubsection{Local Branch}
	
	To achieve local multi-scale feature mapping of the original data, we designed the Local branch, whose structure can be referenced in Fig.\ref{fig:3}(c). In the Local branch, to obtain sufficient multi-scale features during the feature mapping stage, we adopt the classic multi-scale max-pooling (5$\times$5, 9$\times$9, 13$\times$13) on downsampled features. Employing multi-scale max-pooling on downsampled features effectively captures feature information at different scales, thereby enhancing the model's ability to recognize multi-scale features.
	
	Before describing the entire Local branch, we define several operations in advance:
	
	\begin{equation}
	\label{eq:7}
		f_{BL}(x) = \delta_{3\times3}(\delta_{1\times1}(x))
	\end{equation}
	
	\begin{equation}
	\label{eq:8}
		f_{SPP}(x) = Cat({MP}_{5\times5}(x), {MP}_{9\times9}(x), {MP}_{13\times13}(x), x)
	\end{equation}
	
	Here, $f_{BL}(\cdot)$ represents the Bottleneck layer, $f_{SPP}(\cdot)$) denotes the Spatial Pyramid pooling operation, and ${MP}_{k\times k}(\cdot)$ represents max-pooling with a kernel size of $k \times k$, where the outputs and inputs of these two operations are of the same size. For the input features $x$ of the Local branch, $x \in \mathbb{R}^{C\times H\times W}$, we have:
	
	\begin{equation}
	\label{eq:9}
	\begin{split}
	F_m &= f_m(x) \\
	&= \psi(Cat(\delta_{3\times3}(\delta_{3\times3}(x)), \\
	& \quad f_{BL}(f_{SPP}(f_{BL}(\psi(\delta_{3\times3}(x)))))))\\
	\end{split}
	\end{equation}
	
	Here, $F_m \in \mathbb{R}^{2C\times(H/2)\times(W/2)}$ represents the output features of the Local branch.

	\subsubsection{Wavelet Transform Feature Decomposer }
	
	\begin{figure}
		\centering
		\includegraphics[width=0.45\textwidth]{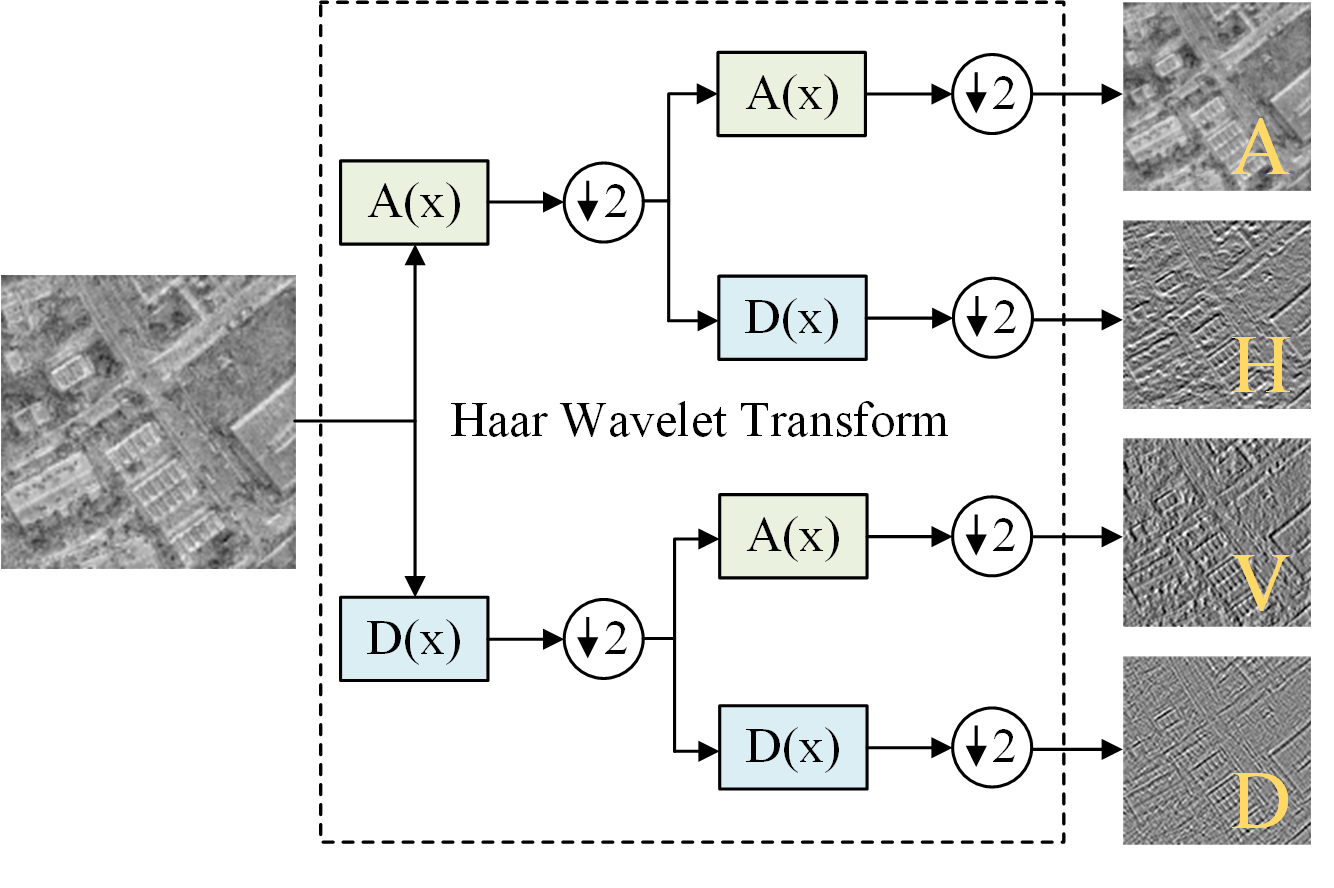}
		\caption{Principle diagram of feature decomposition using Haar Wavelet Transform. Where A(x) represents low-pass filtering of the original data to obtain the low-frequency approximation coefficients, and D(x) represents high-pass filtering of the original data to obtain the high-frequency detail coefficients. Each decomposition reduces the size of the features by half. The proposed WTFD in this paper obtains the low-frequency signal A, horizontal high-frequency signal H, vertical high-frequency signal V, and diagonal high-frequency signal D through two iterations of Haar Wavelet Transform.}
		\label{fig:5}       
	\end{figure}
	
	Images are two-dimensional (or three-dimensional) discrete non-stationary signals that contain information in different frequency ranges. The Haar wavelet transform can perform multi-resolution decomposition of image signals, capturing their local properties and specific adaptability, thus it is widely used to improve the feature representation of image signals, such as denoising, super-resolution, and compression tasks. However, in the field of image segmentation, the application of wavelet transforms is relatively limited. Previous studies usually directly apply wavelet transforms to the downsampling layers of CNN networks, substituting frequency domain features for spatial domain features, but this may lead to spatial information loss. To overcome this issue, we propose a method that simultaneously considers frequency domain and spatial domain information.
	
	\begin{figure*}
		\centering
		\includegraphics[width=0.9\textwidth]{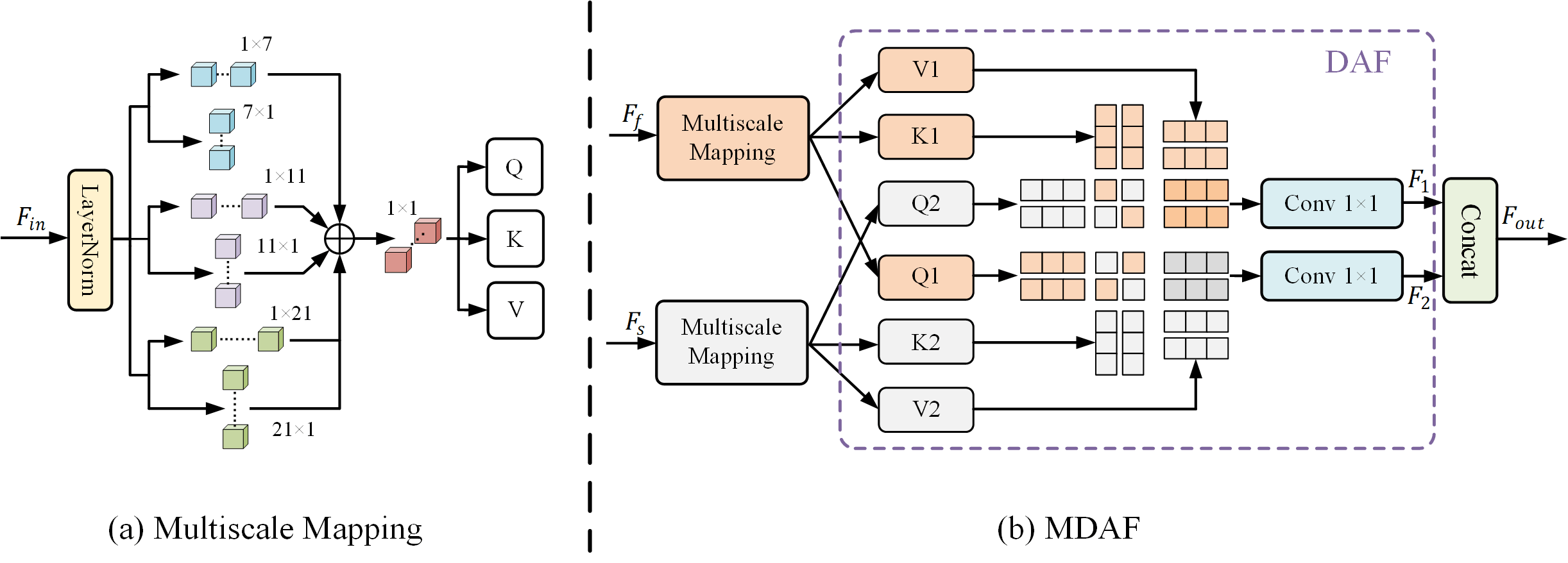}
		\caption{MDAF Structure. (a) Multiscale Mapping. (b) MDAF. This structure is used for semantic alignment and feature selection between frequency domain features and spatial domain features. We first perform multiscale mapping on both types of features to align them to a unified scale. Then, utilizing a Dual-Representation Alignment Filter (DAF) implemented with cross-attention, we facilitate interaction and selection between the two types of features. Finally, the selected features are combined to obtain more representative feature signals for subsequent segmentation.}
		\label{fig:6}       
	\end{figure*}
	
	To introduce frequency domain information into the segmentation network, we design a Wavelet Transform Feature Decomposer (WTFD) module for frequency domain feature mapping.
	Its structure can be seen in Fig.\ref{fig:3}(b).This module utilizes convolution and the Haar wavelet transform to convert spatial domain features into four frequency domain components, one of which is the low-frequency component and the other three are the horizontal, vertical, and diagonal high-frequency components. Then, feature representation learning is performed on the low-frequency information through convolution to obtain low-frequency features. The process of the Haar wavelet transform is illustrated in Fig.\ref{fig:5}, where $A(\cdot)$ represents the low-pass filter, $D(\cdot)$ represents the high-pass filter, and $A$, $H$, $V$, $D$ respectively represent the low-frequency component, horizontal high-frequency component, vertical high-frequency component, and diagonal high-frequency component. The three high-frequency components are combined into a high-frequency signal through concatenation, followed by non-linear operations for feature mapping. In this process, we reduce the channel dimension to filter out redundant information as much as possible, thus obtaining high-frequency features. This method fully utilizes the simple and efficient decomposition of the Haar wavelet transform on image signals, introducing frequency information into the network to achieve a more comprehensive feature representation.
	
	Specifically, for the input $x \in \mathbb{R}^{C\times H\times W}$, a point convolution operation is first performed to increase the non-linearity of the features, resulting in new features $X \in \mathbb{R}^{C\times H\times W}$ with unchanged dimensions. Then, the Haar wavelet transform is used to transform the features into one low-frequency component, one horizontal high-frequency component, one vertical high-frequency component, and one diagonal high-frequency component. For each channel $X_c\in \mathbb{R}^{H\times W}$, the 1st-order Haar wavelet transform is calculated as follows:
	\begin{equation}
	\label{eq:10}
	\left\{ \begin{array}{l}
		A_c(i,j)=\frac{X_c(i,2j-1)+X_c(i,2j)}{2} \\
		\\
		D_c(i,j) = X_c(2i-1,j) - X_c(2i,j) \\
	\end{array} \right. 
	\end{equation}
	
	Where $A_{c(i,j)}$ represents the low-frequency approximation coefficients of channel $ c$, and $D_{c(i,j)}$ represents the high-frequency detail coefficients of channel $c$, where $i$ is the row index and $j$ is the column index. $i$ ranges from 1 to $H$ and $j $ ranges from 1 to $\frac{W}{2}$. Then, the Haar wavelet transform is applied to each column of the approximation coefficients and detail coefficients as follows:
	\begin{equation}
	\label{eq:11}
	\left\{ \begin{array}{l}
	 A_c = AA_c(i,j) = \frac{A_c({i,2j-1})+A_c(i,2j)}{2} \\
	 \\
	 H_c = AD_c(i,j) = \frac{D_c(i,2j-1)-D_c(i,2j)}{2} \\
	 \\
	 V_c = DA_c(i,j) = A_c(i,2j-1)-A_c(i,2j-1) \\
		\\
	 D_c = DD_c(i,j) = D_c(i,2j-1)-D_c(i,2j)\\
	\end{array} \right. 
	\end{equation}
	
	Where $A_c$ represents the approximation coefficients of a single channel, $H_c$ represents the horizontal detail coefficients of a single channel, $V_c$ represents the vertical detail coefficients of a single channel, $D_c$ represents the diagonal detail coefficients of a single channel, and $i$ ranges from 1 to $\frac{H}{2}$ and $j$ ranges from 1 to $\frac{W}{2}$. Finally, the concatenation of the three high-frequency components followed by point convolution for low-dimensional mapping is performed to obtain the final high-frequency and low-frequency features:
	\begin{equation}
	\label{eq:12}
	F_l,F_h = f_w(x) = (\psi(\delta_{1\times1}(A)),\psi(\delta_{1\times1}(Cat(H,V,D))))
	\end{equation}
	
	Where $F_l$ and $F_h$ represent high-frequency and low-frequency features, respectively, with dimensions $(C, H/2, W/2)$, $\psi(\cdot)$ denotes batch normalization calculation, and $\delta_{k\times k}$ represents a convolution kernel size of $k \times k$.
	
	WTFD decomposes spatial domain features into high-frequency features with local details and low-frequency features with global characteristics, providing a more comprehensive feature set for the segmentation network to consider the frequency domain information of features.

	\subsection{Multiscale Dual-Representation Alignment Filter}

	Frequency domain features and spatial domain features capture different aspects and properties of an image, but due to their semantic differences, semantic alignment is required to ensure consistency and complementarity in representing the image. To achieve this semantic alignment and feature selection for both frequency domain and spatial domain features and promote their combination, we designed a Multi-Domain Attention Fusion (MDAF), as shown in Fig.\ref{fig:6}.
	
	Firstly, we use scale mapping to map frequency domain features and spatial domain features to a unified scale. Specifically, we employ vertical bar-shaped convolutions of different scales to process each feature, then concatenate them and use 1$\times$1 convolutions to map them into matrices Q, K, and V of unified scale as the input for the next stage.
	
	Spatial domain features and frequency domain features each obtain two sets of matrices (Q1, K1, V1) and (Q2, K2, V2), then we propose a method called Domain Attention Fusion (DAF), which calculates attention by querying the counterpart and its own key-value pairs, followed by feature weighting, ultimately achieving feature selection. Finally, concatenate the two as the final output feature.
	
	Through DAF, the model can dynamically adjust attention weights based on task requirements and feature attributes, allowing features with similar semantics to receive greater attention and weight, while features with different semantics can be suppressed or ignored. This ensures that the model focuses more on features with similar semantics during feature selection, thereby achieving better semantic alignment.
	
	For the specific derivation, for input $X\in \mathbb{R}^{C\times H\times W\times}$, we define the Multi-scale Ortho-Convolution operation as:
	
	\begin{equation}
	\label{eq:13}
	\begin{split}
	Q,K,V & = f_{mm}(x)\\
	& = \delta_{1\times1}({OC}_7(LN(x))\\
	& \quad +{OC}_{11}(LN(x))+{OC}_{21}(LN(x)))\\
	\end{split}
	\end{equation}
	
	where, ${OC}_k$ represents bar-shaped convolution with kernel size $1\times k$ and $k\times1$, $LN(\cdot)$ represents layer normalization, $f_{mm}(\cdot)$ represents multi-scale mapping operation, and Q, K, V are matrices after using Multi-scale Ortho-Convolution operation, all of which have sizes $(C,H,W)$.
	
	For the entire feature alignment operation, the input spatial domain feature $F_s \in R^{C\times H\times W}$ and frequency domain feature $F_f \in R^{C\times H\times W}$ undergo Multi-scale Ortho-Convolution operations respectively:
	\begin{equation}
	\label{eq:14}
	\left\{ \begin{array}{l}
	Q1,K1,V1=f_{mm}(F_s) \\
	Q1,K1,V1=f_{mm}(F_s) \\
	\end{array} \right. 
	\end{equation}
	
	The attention formula is:
	\begin{equation}
	\label{eq:15}
	Attn(Q,K,V)=Softmax(\frac{QK^T}{\sqrt d})V
	\end{equation}
	
	where $d=C\times H\times W$. Then the steps for DAF calculation are as follows:
	\begin{equation}
	\label{eq:16}
	\left\{ \begin{array}{l}
	F_1=\delta_{1\times 1}(Attn(Q2,K1,V1)) \\
	F_2=\delta_{1\times 1}(Attn(Q1,K2,V2)) \\
	\end{array} \right. 
	\end{equation}
	
	Then MDAF operation can be defined as:
	\begin{equation}
	\label{eq:17}
	F_{out}=f_{mdaf}(F_s,F_f)=Cat(F_1,F_2)
	\end{equation}
	
	where, $F_1$, $F_2$ are the two outputs of DAF, and $F_1, F_2\in \mathbb{R}^{C/2\times H\times W}$, $f_{mdaf}(F_s,F_f)$ represents the use of MDAF operation on spatial domain feature $F_s$ and frequency domain feature $F_f$, and $F_{out}\in \mathbb{R}^{C\times H\times W}$ is the output tensor of $f_{mdaf}$.

	\subsection{Loss function}
	
	The loss function $\mathcal{L}$ used in this paper is a combination of a dice loss $\mathcal{L}_{dice}$ and a cross-entropy loss function $\mathcal{L}_{ce}$, which can be expressed as:
	
	\begin{equation}
	\mathcal{L}_{ce}=-\frac{1}{N}\sum_{n=1}^N{\sum_{k=1}^K{y_{k}^{\left( n \right)}\log \hat{y}_{k}^{\left( n \right)}}} 
	\label{eq:18} 
	\end{equation}
	\begin{equation}
	\mathcal{L}_{dice}=1-\frac{2}{N}\sum_{n=1}^N{\sum_{k=1}^K{\frac{\hat{y}_{k}^{\left( n \right)}y_{k}^{\left( n \right)}}{\hat{y}_{k}^{\left( n \right)}+y_{k}^{\left( n \right)}}}}
	\label{eq:19} 
	\end{equation}
	\begin{equation}
	\mathcal{L}=\mathcal{L}_{ce}+\mathcal{L}_{dice}
	\label{eq:20} 
	\end{equation}
	
	Here, $N$ represents the number of samples, and $K$ represents the number of classes. $ y^{(n)}$ and $\hat{y}^{(n)}$ denote the one-hot encoding of the true semantic labels and their corresponding softmax outputs from the network, where $n\in [1, \dots, N]$. Specifically, $\hat{y}_k^{(n)}$ represents the confidence of sample $n$ belonging to class $k$.
	
	\section{Experiment}
	\subsection{Experimental settings}
	\subsubsection{Datasets}
	The Vaihingen and Potsdam datasets are widely recognized as standard data sources in the field of remote sensing image semantic segmentation. They are commonly used for algorithm performance evaluation to ensure the broad applicability and comparability of research results. These datasets provide diverse land cover categories and environmental conditions, including buildings, roads, trees, etc. They also consider different seasons, weather, and lighting conditions, which help evaluate the robustness and generalization performance of models. Due to the extensive research and use of these datasets, we can easily compare and contrast our work with previous studies. Here is a detailed introduction to these two datasets:
	
	\textbf{Vaihingen Dataset:} This dataset originates from remote sensing images of the Vaihingen area in Germany, comprising 33 high-resolution TOP image blocks, each with an average size of 2494×2064 pixels. The image blocks consist of true orthophoto (TOP), digital surface model (DSM), and normalized digital surface model (NDSM). The dataset covers five foreground land cover classes (impervious surfaces, buildings, low vegetation, trees, cars) and one background land cover class (clutter). In experiments, we strictly selected training data according to specific training IDs provided by the ISPRS competition (IDs 1, 3, 5, 7, 11, 13, 15, 17, 21, 23, 26, 28, 30, 32, 34, 37), and the remaining 17 images were used for testing. This selection ensures consistency with data from other researchers for comparative analysis. We cropped image blocks into small patches of size 1024$\times$1024 pixels for processing.
	
	\textbf{Potsdam Dataset:} This dataset utilizes aerial images from Potsdam, Germany, containing 38 high-resolution TOP image blocks, with a ground sampling distance of 5 centimeters and a size of 6000×6000 pixels per block. Similar to Vaihingen, each image block consists of true TOP and digital surface model (DSM) and provides four multispectral bands (red, green, blue, and near-infrared). During training, we also used specific training IDs provided by the ISPRS competition, including image numbers 2\_10, 2\_11, 2\_12, 3\_10, 3\_11, 3\_12, 4\_10, 4\_11, 4\_12, 5\_10, 5\_11, 5\_12, 6\_7, 6\_8, 6\_9, 6\_10, 6\_11, 6\_12, 7\_7, 7\_8, 7\_9, 7\_11, and 7\_12, while the remaining 15 images were used as the test set. Similarly, we cropped image blocks into 1024$\times$ 1024 pixel patches for analysis. 
	In quantitative evaluation, we excluded the "clutter/background" category.
	\begin{figure}
		\includegraphics[width=0.48\textwidth]{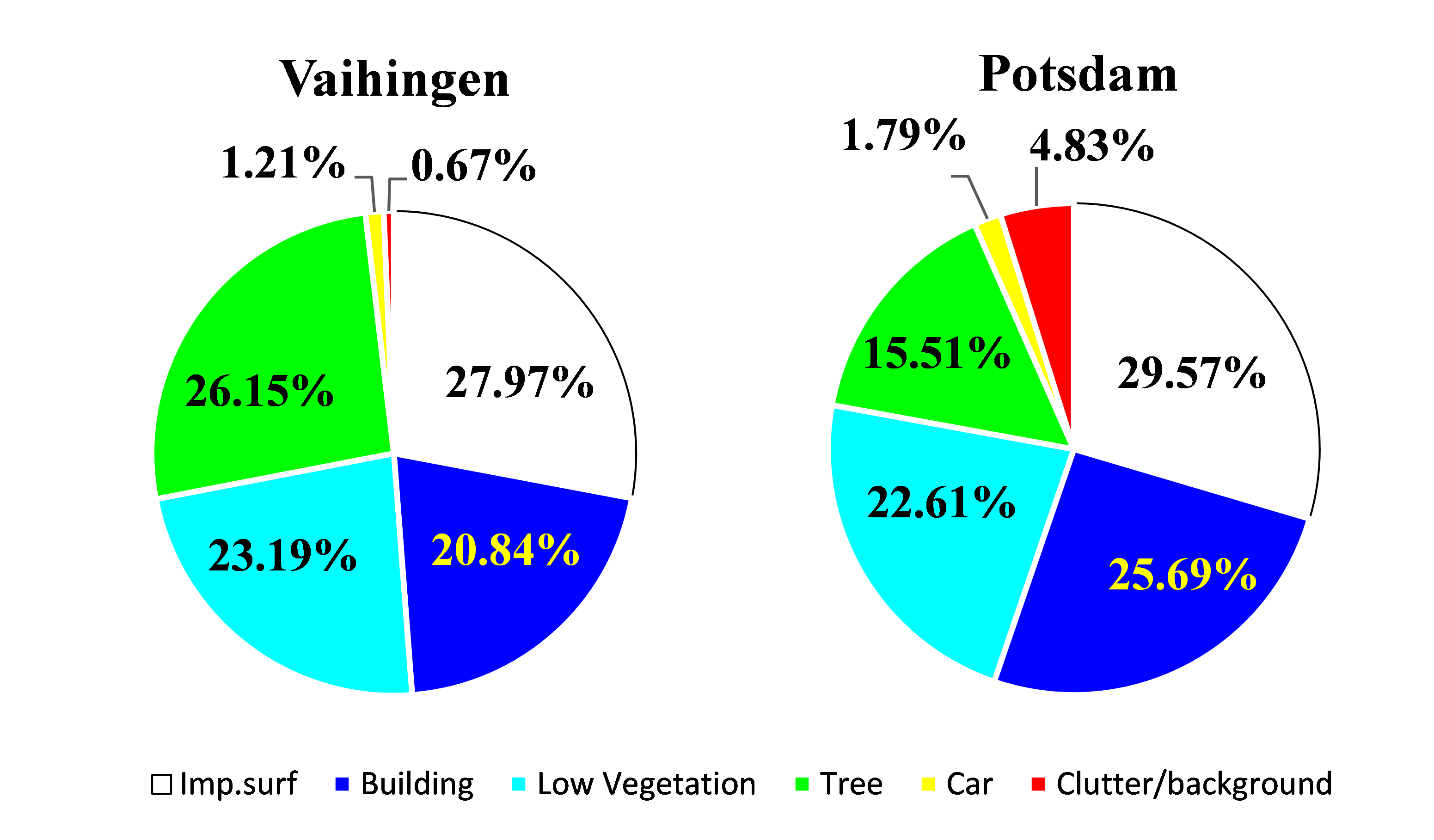}
		\caption{Data set label proportion chart}
		\label{fig:7}       
	\end{figure}

	\subsubsection{Implementation Details}
	
	In this study, we utilized the Ubuntu 18.04 operating system and deployed all models on a single NVIDIA GeForce RTX 2080 Ti 11GB GPU, using the PyTorch 1.11 framework. To accelerate the convergence speed of the models, we employed the AdamW optimizer, with a base learning rate set to 6e-4, and utilized cosine learning rate scheduling. For the Vaihingen and Potsdam datasets, we conducted the following data preprocessing steps: Firstly, we randomly cropped images into patches of size 512×512. During the training phase, we introduced multiple data augmentation techniques, including random scaling (0.5, 0.75, 1.0, 1.25, 1.5), random vertical flipping, random horizontal flipping, and random rotation. The training process consisted of 105 epochs. During the testing phase, we employed multi-scale evaluation and random flipping augmentation techniques to ensure the robustness and performance of the models.

	\subsubsection{Evaluation metrics}
	
	In this experiment, we adopted commonly used metrics in remote sensing segmentation: OA (Overall Accuracy), F1 score, and mIoU (Mean Intersection over Union) as evaluation metrics. Additionally, to assess the model's parameter count, we also used the Parameters and FLOPs metrics. Before introducing these metrics, we need to understand some related terms and symbol meanings: tp (true positives), fp (false positives), fn (false negatives), tn (true negatives).
	
	\textbf{Precision:}Precision measures the proportion of true positive samples among the samples predicted as positive by the model. In simple terms, precision tells us the likelihood that a sample predicted as positive by the model is indeed a true positive.
	
	\begin{equation}
		Precision=\frac{tp}{tp+fp}
		\label{eq:21}
	\end{equation}
	
	\textbf{Recall:} Recall refers to the proportion of true positive samples that the model successfully predicts as positive among all truly positive samples. Recall measures the model's ability to identify all positives.
	
	\begin{equation}
		Recall=\frac{tp}{tp+fn}
		\label{eq:22}
	\end{equation}
	
	\textbf{Overall Accuracy (OA):}OA is one of the commonly used performance evaluation metrics in image classification tasks. It represents the proportion of correctly classified samples to the total number of samples. However, OA may not handle class imbalances well, as the model may tend to predict the class with more samples when some classes have far more samples than others.
	
	\begin{equation}
	 	OA=\frac{tp+tn}{tp+fp+fn+tn}
	 	\label{eq:23}
	\end{equation}
	
	\textbf{F1 Score:} The F1 score is the harmonic mean of precision and recall. It comprehensively considers the model's accuracy and its ability to capture positives. For multi-class problems, F1 scores are usually calculated for each class and then averaged.
	\begin{equation}
		F1=\frac{2\times \left( Precision\times \text{Re}call \right)}{Precision+\text{Re}call}
		\label{eq:24}
	\end{equation}
	Overall F1 Score = Average of F1 Scores for all classes.
	
	\textbf{Mean Intersection over Union (mIoU):} mIoU is a commonly used evaluation metric in semantic segmentation tasks, measuring the accuracy of the model in pixel-level segmentation. IoU (Intersection over Union) is used to measure the segmentation results for each class, while mIoU computes the average IoU for all classes.
	\begin{equation}
		IoU=\frac{tp}{tp+fp+fn}
		\label{eq:32}
	\end{equation}
	mIoU is the sum of IoU values for all categories divided by the number of categories.

	\subsection{Ablation Experiment}
	To assess the performance of different components of SFFNet, we conducted a series of ablation experiments, validated using the Vaihingen and Potsdam datasets. In the discussion, we primarily focus on two performance metrics: mIoU (mean Intersection over Union) and meanF1. All experimental results are averages obtained from multiple trials.
	\subsubsection{Components of SFFNet}
	
	Tab.\ref{tab:1} presents the results of SFFNet with individual modules removed. Here, "Global" denotes the global branch, "Local" refers to the local branch, "WTFD-L" represents the low-frequency part decomposed by WTFD, "WTFD-H" represents the high-frequency part decomposed by WTFD, and "MDAF" stands for Multiscale Dual-Representation Alignment Filter (MDAF). To validate the effectiveness of MDAF, we also conducted comparisons by removing MDAF and directly concatenating (marked as "Cat") or adding (marked as "Add") frequency domain features and spatial domain features. For experiments where either the low-frequency or high-frequency part is removed from SFFNet, the corresponding feature alignment by MDAF is also removed simultaneously. The components removed in the ablation experiments are indicated by "(w/o)," and the added components are marked with "(+)."
	
	\begin{table}
		\centering
		\begin{threeparttable}
			\renewcommand\arraystretch{1.5}
			\centering
			\caption{Results of SFFNet with individual components removed.}
			\setlength{\tabcolsep}{1.5mm}{
				\begin{tabular}{l|cc|ccc}
					\toprule
					\midrule
					\multicolumn{1}{c|}{\multirow{2}{*}{\textbf{Method}}} & \multicolumn{2}{c|}{\textbf{Vaihingen}} & \multicolumn{2}{c}{\textbf{Potsdam}} \\\cline{2-5} 
					& \textbf{mIoU(\%)}       &\textbf{F1(\%)}       & \textbf{mIoU(\%)}      & \textbf{F1(\%)}      \\
					\midrule
					SFFNet & 84.80 & 91.67 & 87.73 & 93.36 \\
					SFFNet w/o Global & 83.73 & 91.04 & 86.50 & 92.65 \\
					SFFNet w/o Local & 83.75 & 91.05 & 86.68 & 92.76 \\
					SFFNet w/o WTFD-L & 83.99 & 91.19 & 86.55 & 92.68 \\
					SFFNet w/o WTFD-H & 84.11 & 91.26 & 86.93 & 92.90 \\
					SFFNet w/o MDAF + Cat & 84.21 & 91.31 & 87.17 & 93.04 \\
					SFFNet w/o MDAF + Add & 84.02 & 91.21 & 87.01 & 92.93 \\
					\midrule
					\bottomrule
			\end{tabular}}
			\label{tab:1}
		\end{threeparttable}
		
	\end{table}
	
	\begin{table}
		\centering
		\begin{threeparttable}
			\renewcommand\arraystretch{1.7}
			\caption{Results of Adding Individual Modules on Baseline Model on the Vaihingen Dataset}
			\centering
			\setlength{\tabcolsep}{5.4mm}{
				\begin{tabular}{l|cc}
					\toprule
					\midrule
					\multicolumn{1}{c|}{\textbf{Method}} & \textbf{mIoU(\%)} & \textbf{F1(\%)} \\
					\midrule
					Baseline & 81.70 & 89.78 \\
					Baseline + Global & 82.61 & 90.36 \\
					Baseline + Local & 82.73 & 90.41 \\
					Baseline + WTFD-H & 82.29 & 90.15 \\
					Baseline + WTFD-L & 82.88 & 90.51 \\
					\midrule
					\bottomrule
			\end{tabular}}
			\label{tab:2}
		\end{threeparttable}

	\end{table}
	
	\begin{table}
		\centering
		\begin{threeparttable}
			\renewcommand\arraystretch{1.5}
			\centering
			\caption{Comparison of parameter count and Flops of different global feature mapping modules, as well as mIoU and F1 scores on the Vaihingen dataset.}
			\setlength{\tabcolsep}{1mm}{
				\begin{tabular}{l|cc|cc}
					\toprule
					\midrule
					\multicolumn{1}{c|}{\multirow{1}{*}{\textbf{Method}}} & \multicolumn{1}{c}{\textbf{Params(M)$\downarrow$}} & \multicolumn{1}{c|}{\textbf{Flops(G)$\downarrow$}} & \multicolumn{1}{c}{\textbf{mIoU(\%)$\uparrow$}}&
					\multicolumn{1}{c}{\textbf{F1(\%)$\uparrow$}}\\
					\midrule
					Vit-Block & 3.68 & 3.82 & 84.38 & 91.13 \\
					Swin-T-Block & 0.40 & 6.58 & 84.72 & 91.57 \\
					Mobile-Vit-Block & 0.51 & 8.42 & 84.39 & 91.32 \\
					Fast-Vit-Block & \textbf{0.22} & 3.55 & 84.59 & 91.41 \\
					\midrule
					\textbf{Global(Ours)} & 0.48 & \textbf{1.99} & \textbf{84.80} & \textbf{91.67} \\
					\midrule
					\bottomrule
			\end{tabular}}
			\begin{tablenotes} 
				\item The best results are indicated in bold black.In the table header, an upward arrow indicates that a higher evaluation value is better, and a downward arrow indicates that a higher evaluation value is better. 
			\end{tablenotes} 
			\label{tab:3}
		\end{threeparttable}
		
	\end{table}
	
	\begin{table}[htbp]
		\centering
		\begin{threeparttable}
			
			\renewcommand\arraystretch{1.5}
			\centering
			\caption{ Number of parameters for different backbones on the Vaihingen dataset's mIoU}
			\setlength{\tabcolsep}{1.5mm}{
				\begin{tabular}{l|c|cc}
					\toprule
					\toprule
					\multicolumn{1}{c|}{\textbf{Method}} & \textbf{Backbone} & \textbf{Parameters(M)$\downarrow$} & \textbf{mIoU(\%) $\uparrow$} \\
					\midrule
					\multicolumn{1}{l|}{\multirow{4}{*}{SFFNet}} & ResNet50     & 25.55 & 84.41 \\
					& ResNext50     & 25.01 & 84.48 \\
					& ResNest50     & 27.46 & 84.54 \\
					& ConvNext-Tiny & 28.57 & 84.80 \\
					\bottomrule
					\bottomrule
			\end{tabular}}
			\label{tab:4}%
		\end{threeparttable}
	\end{table}%

	\begin{figure}
		\centering
		\includegraphics[width=0.5\textwidth]{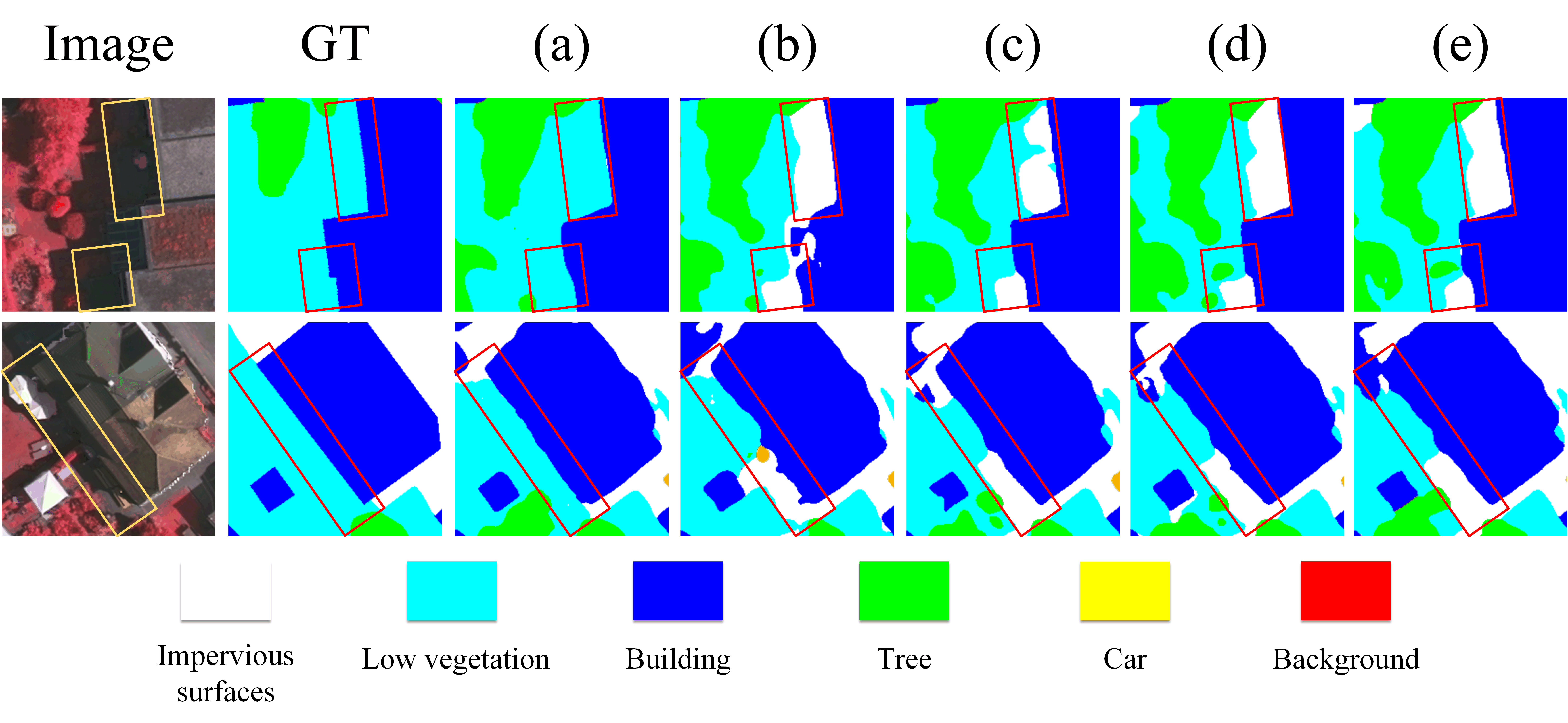}
		\caption{This Figure shows the locally enlarged segmentation results of removing various components in SFFNet. (a) represents SFFNet. (b) represents SFFNet without the global branch. (c) represents SFFNet without the local branch. (d) represents SFFNet without WTFD-L. (e) represents SFFNet without WTFD-H. It can be observed from the figure that removing any branch results in a decline in segmentation performance.}
		\label{fig:8}       
	\end{figure}
	\begin{figure}
		\centering
		\includegraphics[width=0.46\textwidth]{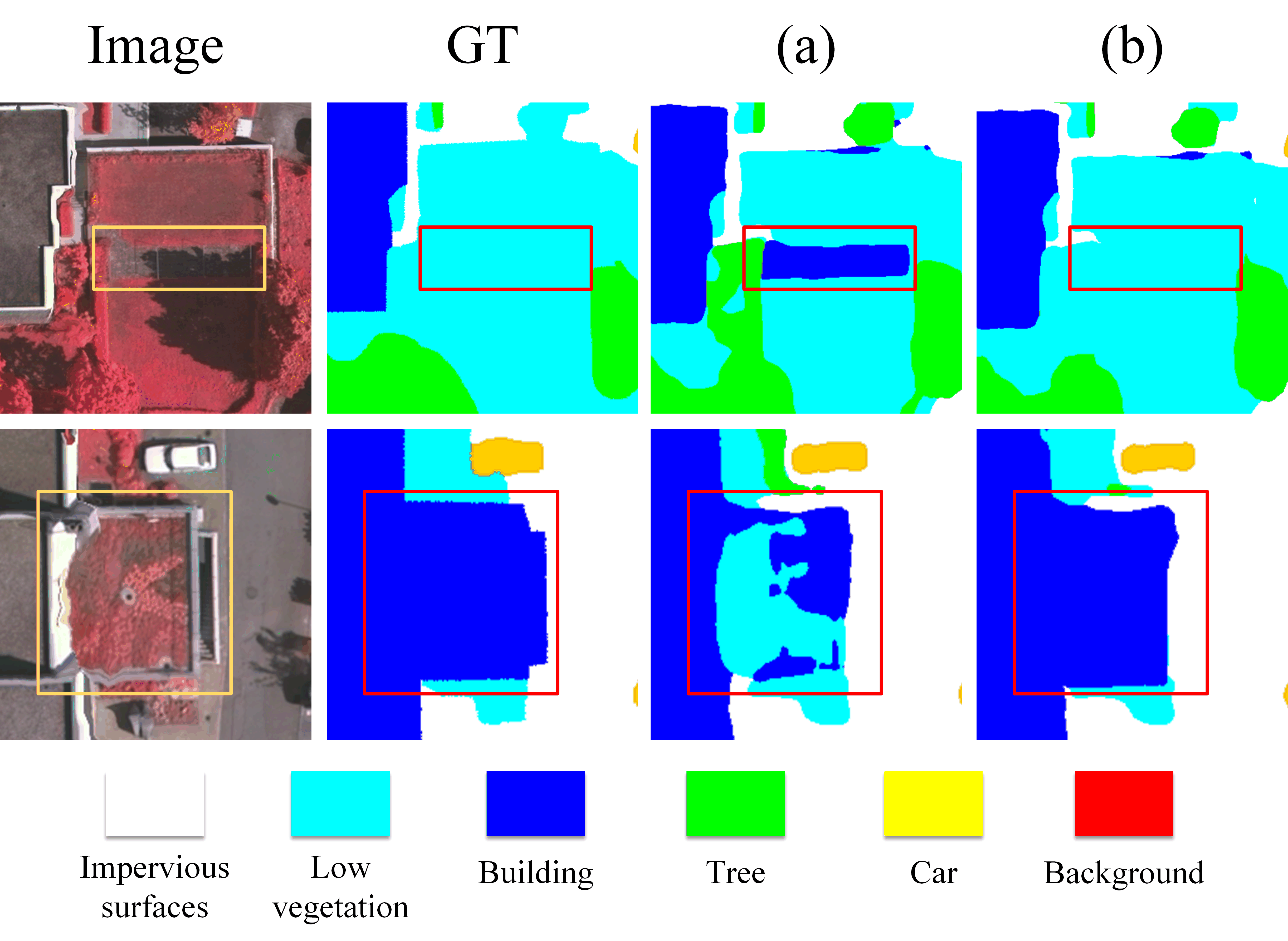}
		\caption{Local enlarged segmentation results before and after adding the Global branch in the baseline. (a) corresponds to the baseline. (b) corresponds to Baseline+GLB. It can be observed from the images that the model shows improved performance in segmenting large continuous regions after adding the Global branch.}
		\label{fig:9}       
	\end{figure}

	\begin{figure}
		\centering
		\includegraphics[width=0.46\textwidth]{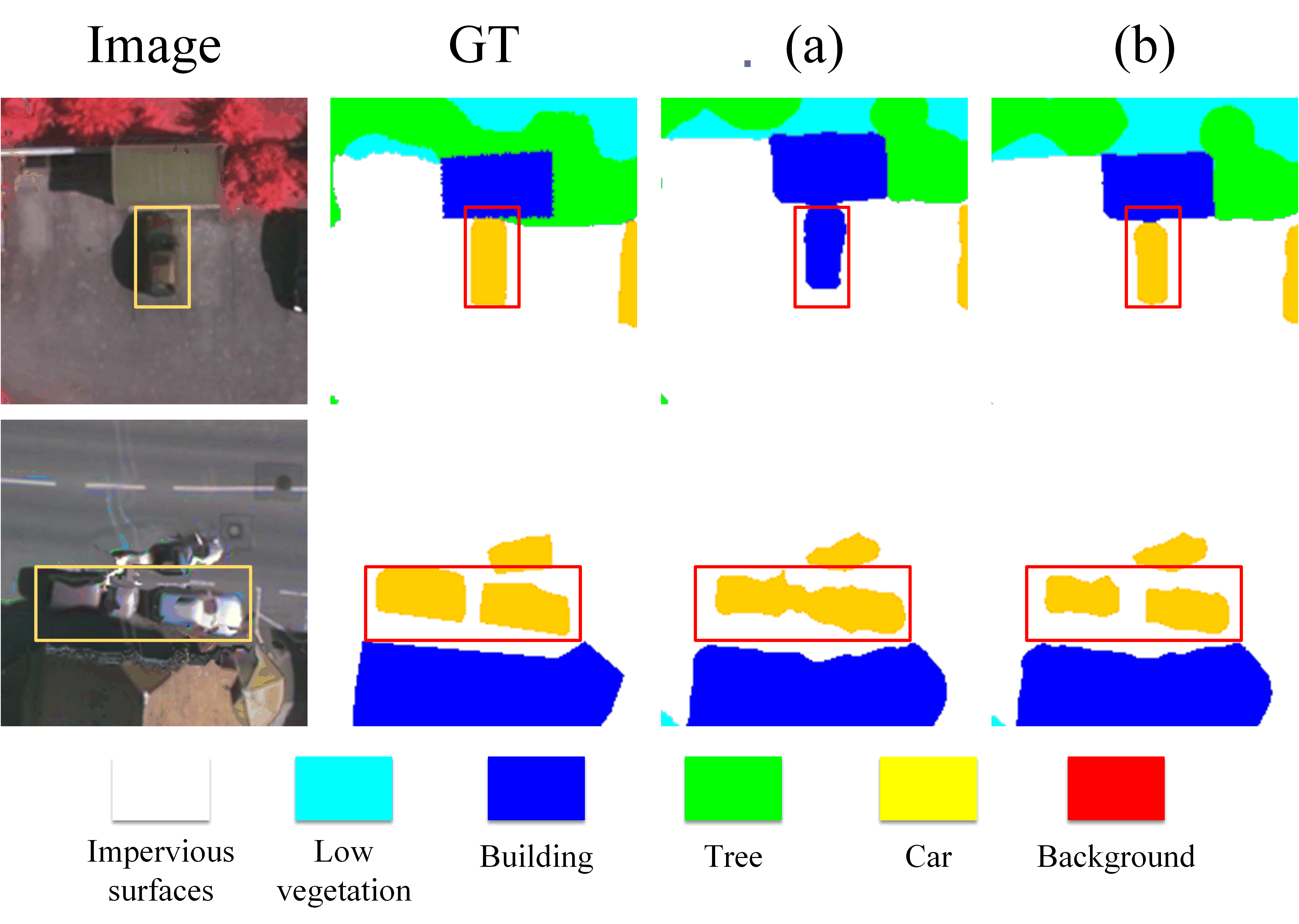}
		\caption{Local enlarged segmentation results before and after adding the Local branch in the baseline. (a) corresponds to the baseline. (b) corresponds to Baseline+Local. The images demonstrate that the model using the Local branch achieves better segmentation results for local details.}
		\label{fig:10}       
	\end{figure}
	
	\begin{figure}
		\centering
		\includegraphics[width=0.46\textwidth]{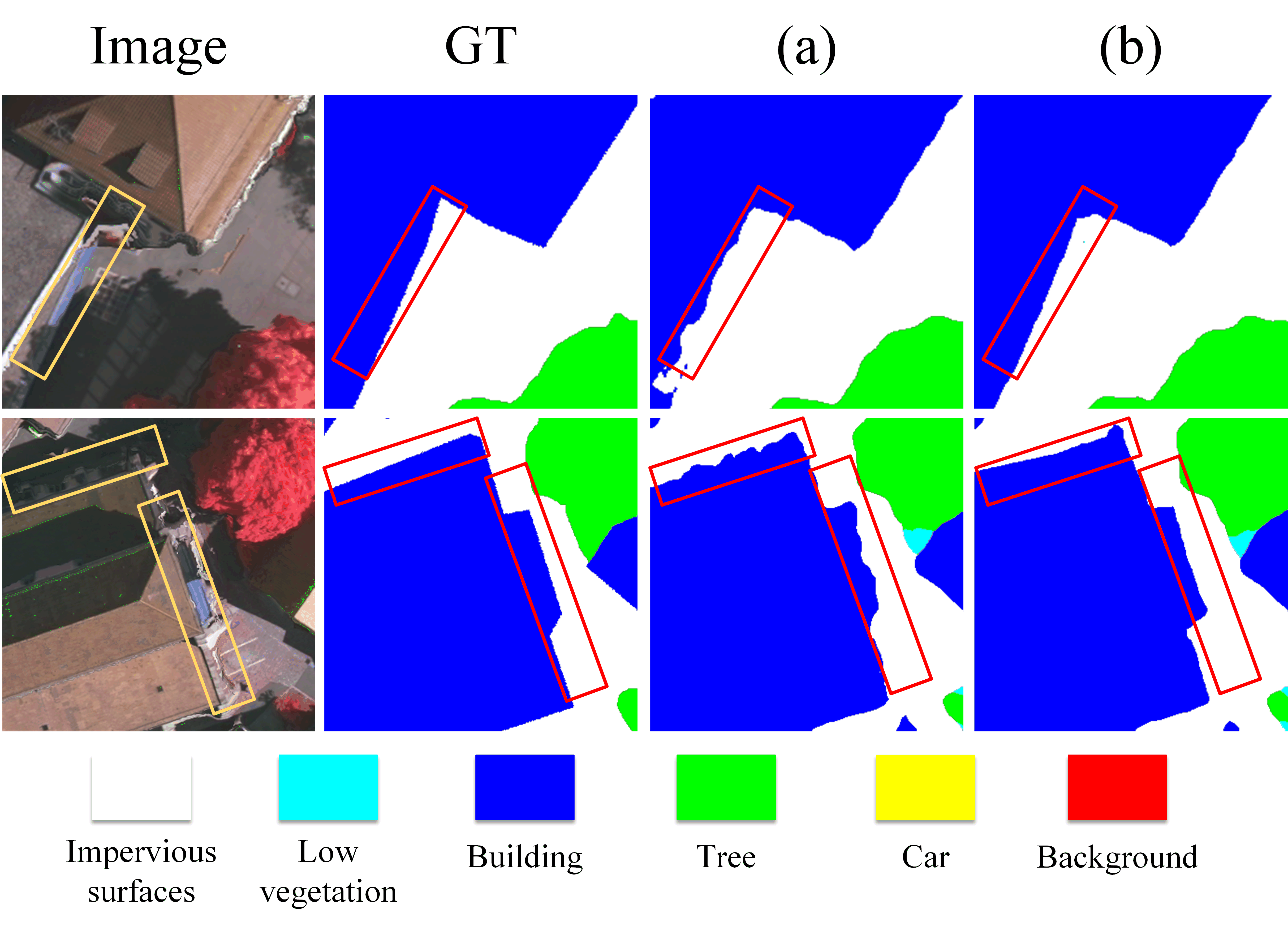}
		\caption{Local enlarged segmentation results before and after adding low-frequency features in the baseline. (a) corresponds to the baseline. (b) corresponds to Baseline+WTFD-L. It can be seen that using low-frequency features has a noticeable effect on objects difficult to segment under shadows.}
		\label{fig:11}       
	\end{figure}
	
	\begin{figure}
		\centering
		\includegraphics[width=0.46\textwidth]{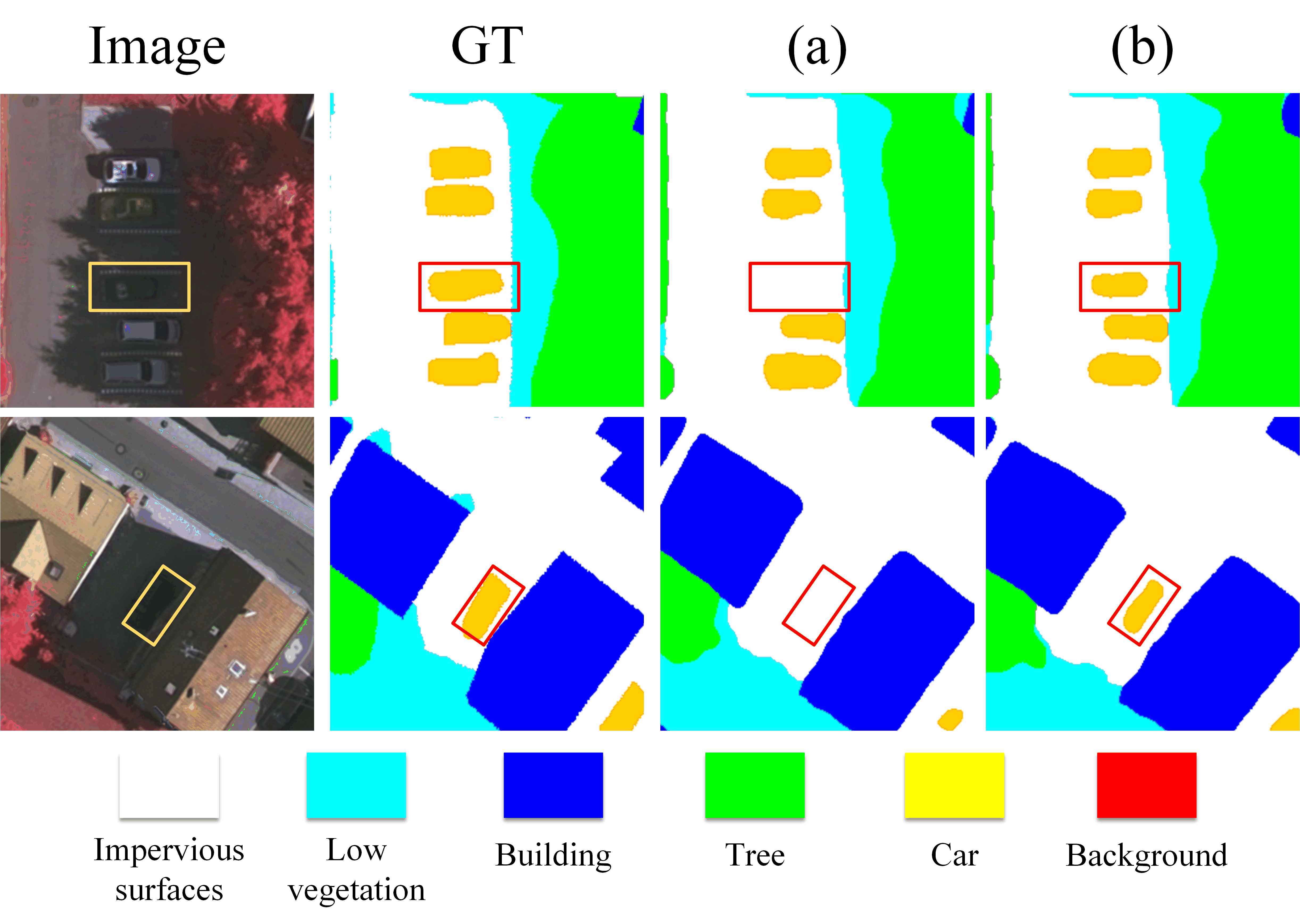}
		\caption{Local enlarged segmentation results before and after adding high-frequency features in the baseline. (a) corresponds to the baseline. (b) corresponds to Baseline+WTFD-H. Improvement in the edge parts is observed after adding high-frequency features to the model.}
		\label{fig:12}       
	\end{figure}
	\begin{figure}
		\centering
		\includegraphics[width=0.46\textwidth]{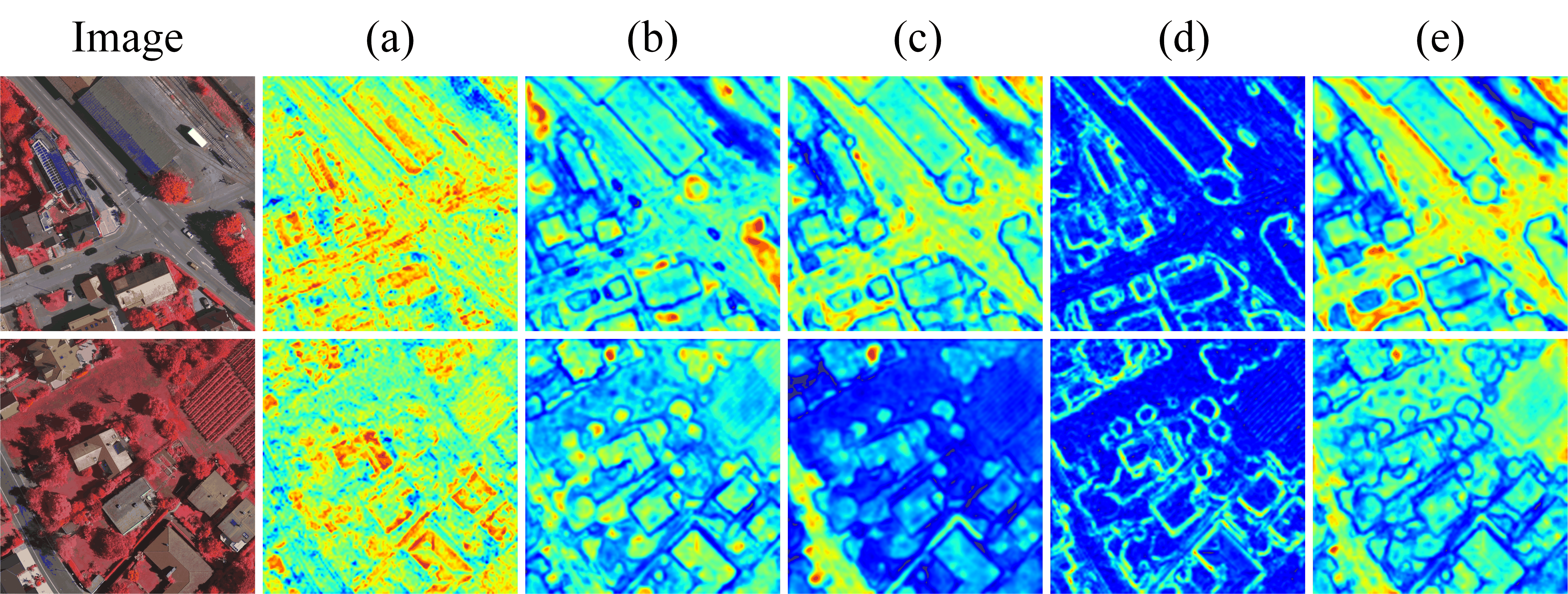}
		\caption{The effects of four types of feature mappings on the extracted features in the first stage. (a) shows the original features before mapping. (b) shows the features after local mapping. (c) shows the features after global mapping. (d) displays the decomposed high-frequency features, and (e) shows the decomposed low-frequency features. It can be observed that local features focus more on local details, such as small objects, while global features focus on large-scale continuous regions. High-frequency features primarily focus on local edges, and low-frequency features also emphasize large-scale continuous regions, but they tend to pay more attention to shadowed areas and regions with multiple textures compared to spatial global features. However, spatial features lack attention to shadowed areas and regions with multiple textures.}
		\label{fig:13}       
	\end{figure}
	\begin{figure}
		\centering
		\includegraphics[width=0.5\textwidth]{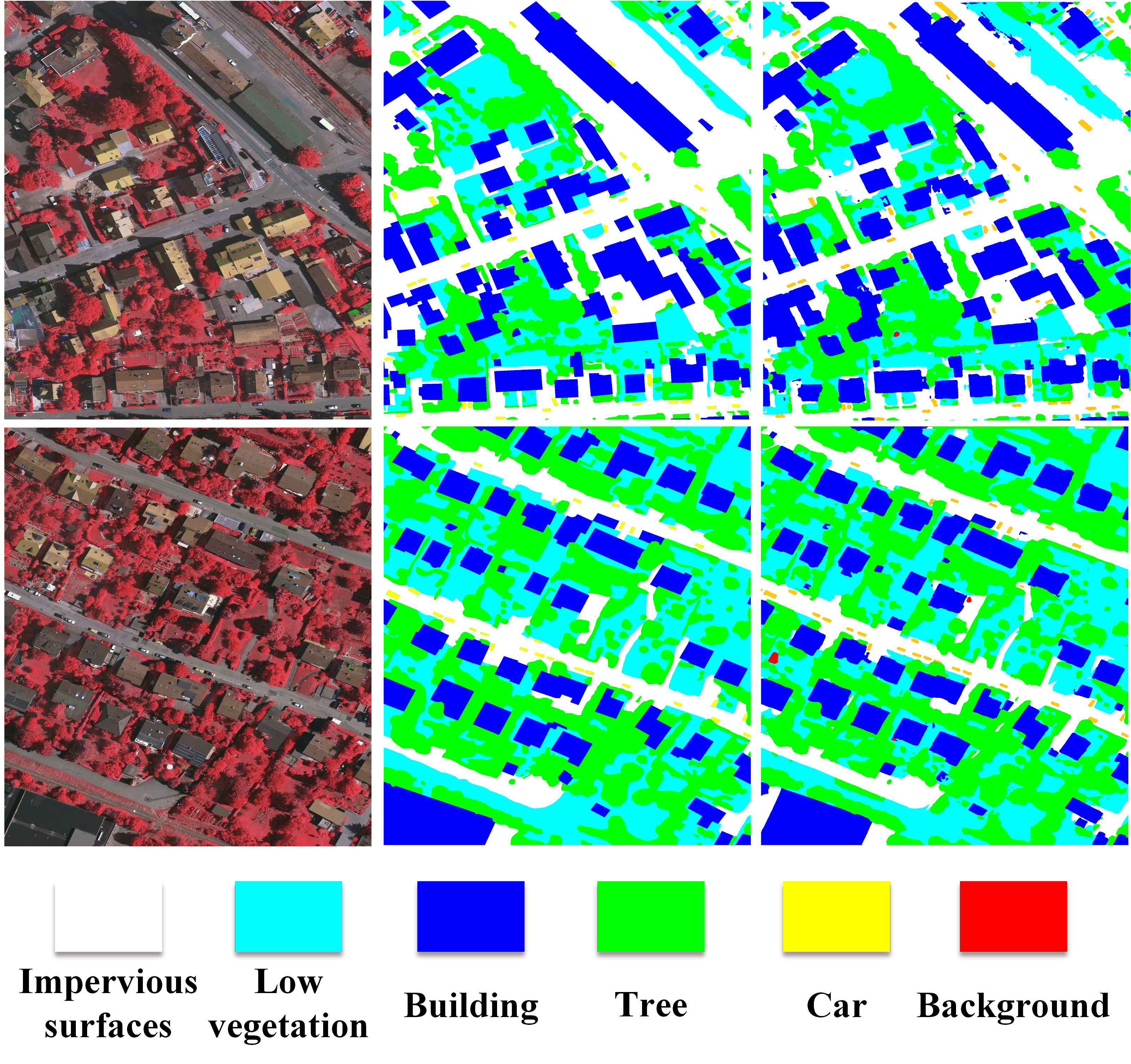}
		\caption{Segmentation results of IDs 10 and 12 on the Vaihingen dataset. The first column shows the original image, the second column shows the ground truth (GT), and the third column shows the segmentation results of SFFNet.}
		\label{fig:14}       
	\end{figure}
	
	\begin{figure*}
		\centering
		\includegraphics[width=0.9\textwidth]{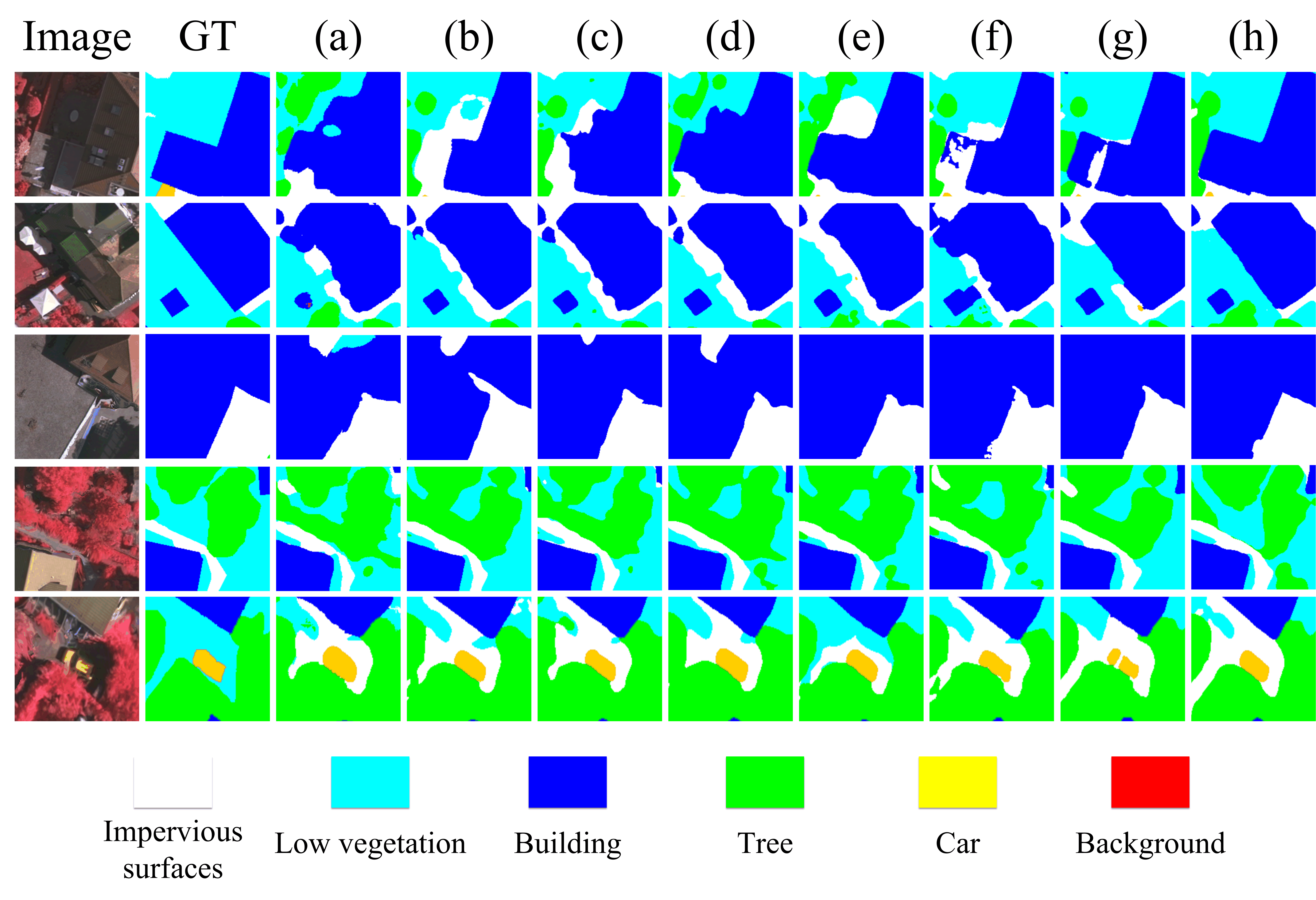}
		\caption{Local enlargement of segmentation results on the Vaihingen dataset for SFFNet and other comparative models. (a) Corresponds to ABCNet. (b) Corresponds to MAResU-Net. (c) Corresponds to A2-FPN. (d) Corresponds to ST-Unet. (e) Corresponds to DC-Swin. (f) Corresponds to MPCNet. (g) Corresponds to Xnet. (h) Corresponds to SFFNet (Ours). From the figure, it can be observed that our model performs better in segmenting areas with large grayscale variations compared to spatial segmentation methods, while also having more comprehensive spatial information compared to purely frequency domain segmentation methods.}
		\label{fig:15}       
	\end{figure*}
	\begin{figure}
		\centering
		\includegraphics[width=0.5\textwidth]{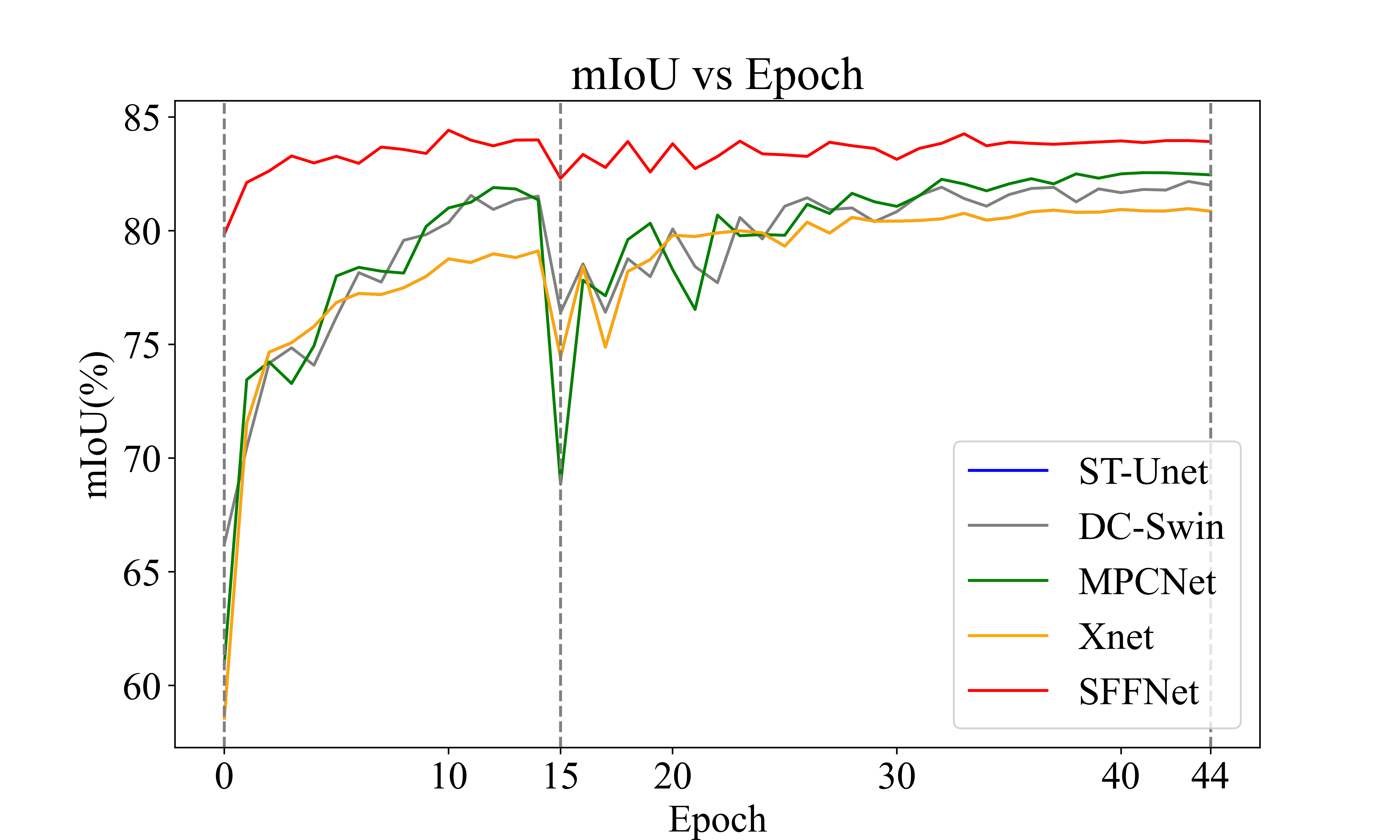}
		\caption{mIoU variation of SFFNet and several mainstream segmentation methods with Epoch. From the figure, it can be seen that SFFNet converges faster and is more stable compared to other networks.}
		\label{fig:16}       
	\end{figure}
	\begin{figure}
		\includegraphics[width=0.51\textwidth]{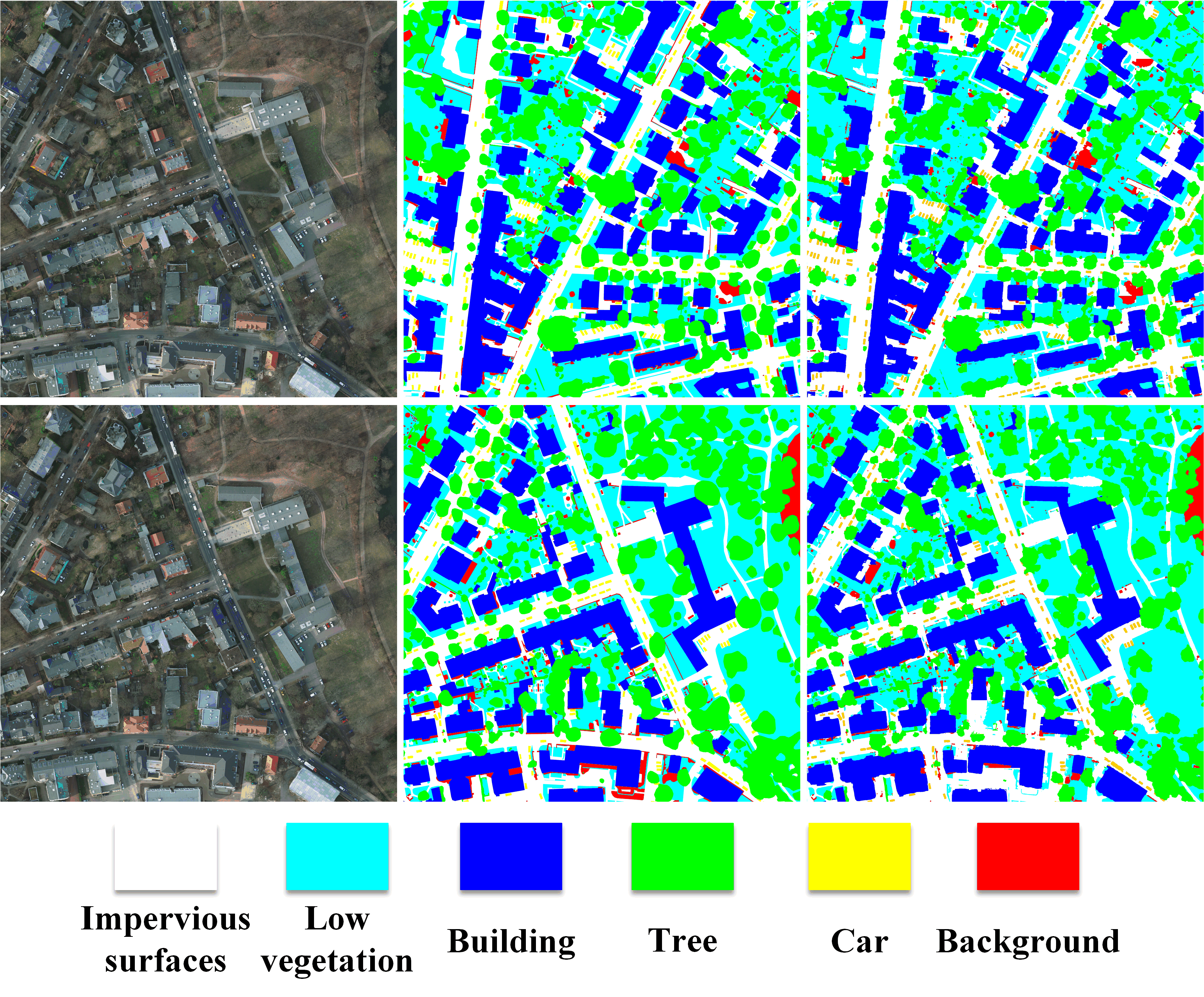}
		\caption{Segmentation results of IDs 3\_13 and 3\_14 on the Vaihingen dataset. The first column shows the original image, the second column shows the ground truth (GT), and the third column displays the segmentation results of SFFNet.}
		\label{fig:17}       
	\end{figure}
	\begin{figure*}
		\centering
		\includegraphics[width=0.9\textwidth]{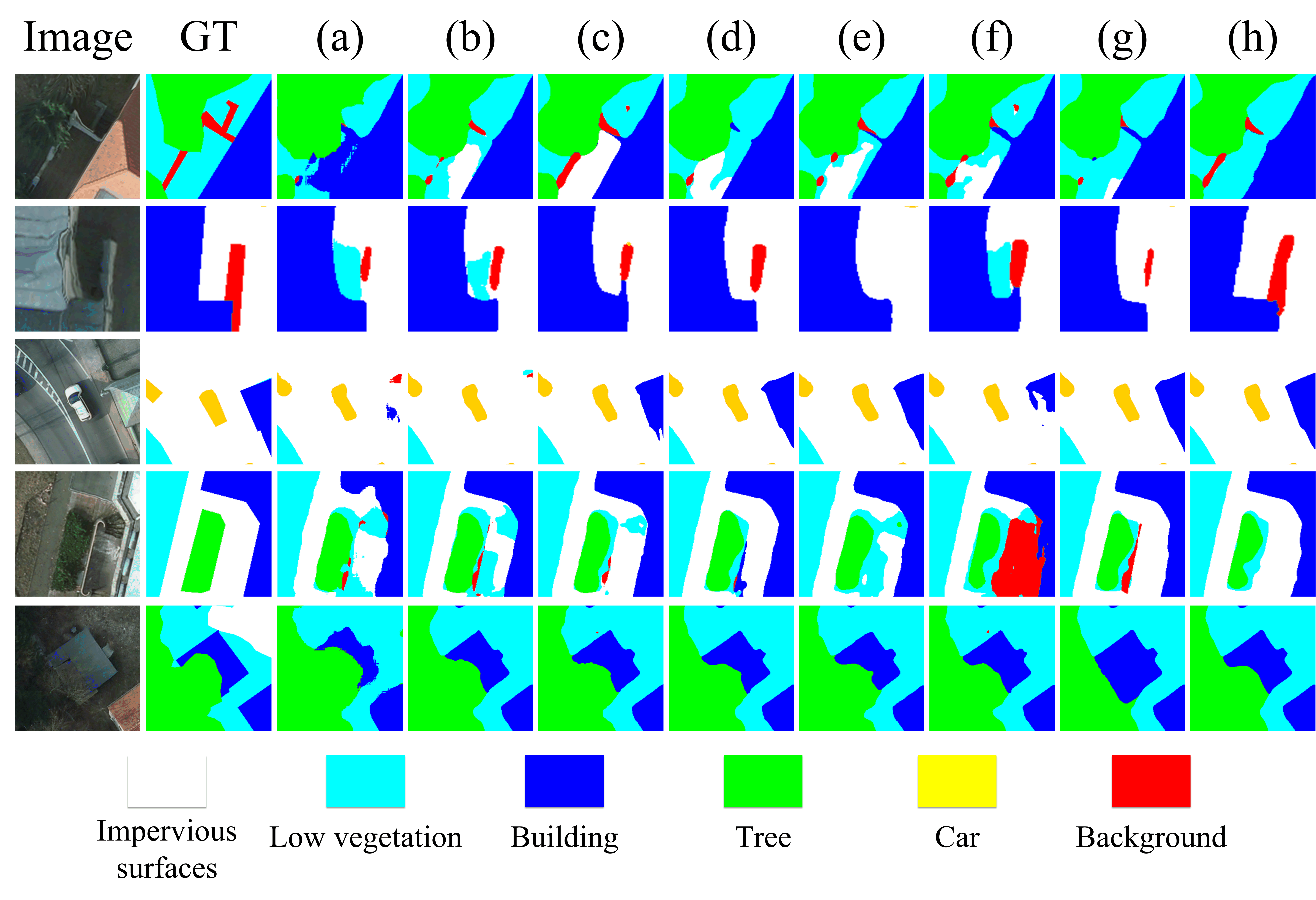}
		\caption{Local enlarged view of segmentation results of SFFNet and other comparative models on the Potsdam dataset. (a) corresponds to ABCNet. (b) corresponds to MAResU-Net. (c) corresponds to A2-FPN. (d) corresponds to ST-Unet. (e) corresponds to DC-Swin. (f) corresponds to MPCNet. (g) corresponds to Xnet. (h) corresponds to SFFNet (Ours). Similar to the results on the Vaihingen dataset, it can be observed from the figure that our model performs better in segmenting areas with large grayscale variations compared to spatial domain segmentation methods, while compared to purely frequency domain segmentation methods, SFFNet also preserves more comprehensive spatial information.}
		\label{fig:18}       
	\end{figure*}

	Fig.\ref{fig:7} shows the segmentation results after removing individual modules from SFFNet. Observing the results, it is evident that the segmentation performance of SFFNet is weakened when any module is removed. To further validate the effectiveness of each component, we conducted experiments on the Vaihingen dataset by incrementally adding individual modules to the baseline model. We set the baseline model as directly obtaining segmentation results through a segmentation head after the spatial features of the first stage of SFFNet, with ConvNext used as the encoder. Tab.\ref{tab:2} presents the experimental results of adding individual modules to the baseline model. Fig.\ref{fig:8}, Fig.\ref{fig:9}, Fig.\ref{fig:10} and Fig.\ref{fig:11} respectively demonstrate the effects of adding any single component to the baseline model, while Fig.\ref{fig:12} illustrates the four different feature maps (local features, global features, high-frequency features, and low-frequency features) after mapping in the second stage. Overall, the comprehensive results indicate that each component proposed in our study positively impacts the model performance.

	\subsubsection{Impact of Global branch}
	
	According to the data in Tab.\ref{tab:1}, after removing the Global branch in the Vaihingen dataset, the mIoU and F1 of SFFNet decreased by 1.07\% and 0.63\%, respectively. Similarly, in the Potsdam dataset, removing the Global branch led to a decrease in mIoU and F1 by 1.23\% and 0.71\%, respectively. From Tab.\ref{tab:2}, it can be observed that adding the Global branch to the baseline model resulted in a performance improvement of 0.91\% in mIoU and 0.58\% in F1.
	
	Visualization results in Fig.\ref{fig:9} show that adding the Global branch to the baseline model improves the segmentation of large continuous objects. Additionally, from Fig.\ref{fig:13}(c), it can be observed that features using the Global branch have higher discriminability in large continuous regions, further validating the Global branch's capability of global modeling.
	
	To verify the improved efficiency of the enhanced Global branch in global mapping, we compared it with other mainstream modules with global mapping capabilities, including Vit\cite{vit}, Swin-T\cite{swintransformer}, Mobile-Vit\cite{mobilevit}, and Fast-Vit\cite{fastvit}. Experimental results are shown in Tab.\ref{tab:3}, indicating that our Global branch performs best in terms of FLOPs and segmentation accuracy, further demonstrating the model's ability to perform global mapping more efficiently.

	\subsubsection{Impact of Local branch}
	
	According to the results in Tab.\ref{tab:1}, removing the Local branch for SFFNet on the Vaihingen dataset resulted in a decrease of 1.05\% in mIoU and 0.62\% in F1. Similarly, on the Potsdam dataset, removing the Local branch led to a decrease of 1.05\% in mIoU and 0.6\% in F1. From Tab.\ref{tab:2}, it can be seen that adding the Local branch to the baseline model increased mIoU by 1.03\% and F1 by 0.63\%.
	
	Through visualization results in Fig.\ref{fig:10}, it is observed that adding the Local branch to the baseline model improves the segmentation of small local objects. Moreover, from Fig.\ref{fig:13}(b), it can be seen that features using the Local branch can better distinguish local details, further validating the Local branch's ability to model local information.

	\subsubsection{Impact of WTFD}
	
	Initially, the original features were decomposed into high-frequency and low-frequency features by WTFD, which contain information lacking in spatial features. According to the data in Tab.\ref{tab:1}, removing the low-frequency features for SFFNet on the Vaihingen dataset resulted in a decrease of 0.81\% in mIoU and 0.48\% in F1. Similarly, on the Potsdam dataset, removing the low-frequency features led to a decrease of 1.18\% in mIoU and 0.68\% in F1. Conversely, adding low-frequency features to the baseline model increased mIoU by 1.18\% and F1 by 0.73\%. On the other hand, according to Tab.\ref{tab:1}, removing the high-frequency features for SFFNet on the Vaihingen dataset resulted in a decrease of 0.69\% in mIoU and 0.41\% in F1. Similarly, on the Potsdam dataset, removing the high-frequency features led to a decrease of 0.8\% in mIoU and 0.46\% in F1. Conversely, adding high-frequency features to the baseline model increased mIoU by 0.59\% and F1 by 0.37\%.
	
	Through visualization results in Fig.\ref{fig:11}, it is observed that adding low-frequency features to the baseline model improves the segmentation of objects with large shaded areas. From Fig.\ref{fig:13}(e), it can be seen that low-frequency features and global features have good discriminability for continuous large objects, and they also have high attention in large continuous shadow areas, which are lacking in spatial features. On the other hand, through visualization results in Fig.\ref{fig:12}, it is observed that adding high-frequency features to the baseline model improves the segmentation of edge areas. From Fig.\ref{fig:13}(d), it can be seen that high-frequency features focus more on local edge information, which is lacking in spatial local features. The introduction of frequency domain features through WTFD significantly improves the segmentation results of the model in areas with significant grayscale variations.

	\subsubsection{Impact of MDAF}
	
	From Tab.\ref{tab:1}, it can be observed that replacing MDAF with Concat resulted in a decrease of 0.59\% and 0.36\% in mIoU and F1 scores, respectively, on the Vaihingen dataset, and a decrease of 0.56\% and 0.32\% on the Potsdam dataset. Conversely, replacing MDAF with Addition led to a decrease of 0.78\% and 0.46\% in mIoU and F1 scores, respectively, on the Vaihingen dataset, and a decrease of 0.72\% and 0.43\% on the Potsdam dataset. These reductions in performance are attributed to the semantic differences between frequency domain and spatial domain features, as direct Concat and Addition without achieving semantic alignment fused the features from the two representation domains. In contrast, our MDAF first aligns the features from the two representation domains before fusion, leading to better performance as evidenced by the experimental results.
	
	\subsubsection{Impact of Backbone}
	
	In order to eliminate the influence of backbone on the results, we conducted a series of replacement experiments on the Vaihingen dataset, using several commonly used and widely adopted backbone models, including ConvNext-Tiny, ResNet50, ResNext50, and ResNest50. The experimental results are shown in Tab.\ref{tab:4}. Observing the data in the table, it can be seen that although the ConvNext-Tiny model has a larger number of parameters, the SFFNet model performs best when using ConvNext-Tiny as the backbone. Therefore, we are more inclined to consider ConvNext-Tiny as the preferred backbone model.

	\begin{table*}[htbp]
		\centering
		\begin{threeparttable}
			\renewcommand\arraystretch{1.7}
			\centering
			\caption{Comparison of Segmentation Results on the Vaihingen Dataset.}
			\setlength{\tabcolsep}{4mm}{
				\begin{tabular}{l|ccccc|cccc}
					\toprule
					\toprule
					\multicolumn{1}{c|}{\multirow{2}{*}{\textbf{Method}}} & \multicolumn{5}{c|}{\textbf{F1(\%)}}                    & \multicolumn{3}{c}{\textbf{Evaluation   index}} \\\cline{2-5} \cline{6-9} 
					& \textbf{Imp.Surf} & \textbf{Building} & \textbf{Lowveg} & \textbf{Tree}  & \textbf{Car}   & \textbf{MeanF1(\%) }   & \textbf{mIoU(\%)}    & \textbf{OA(\%)}   \\
					\midrule
					ABCNet\cite{abcnet}          & 88.13    & 90.23    & 76.71   & 87.21 & 68.72 & 82.20          & 70.58       & 85.62    \\
					MAResU-Net\cite{resunet}      & 92.91    & 95.26    & 84.95   & 89.94 & 88.33 & 90.28         & 83.30        & 90.86    \\
					A2-FPN\cite{a2fpn}          & 92.99    & 95.53    & 84.67   & 90.34 & 87.62 & 90.23         & 82.42       & 91.04    \\
					ST-Unet\cite{stunet} & 92.00    & 94.85    & 83.87   & 90.88 & 86.40 & 89.60        & 81.39       & 90.53   \\
					DC-Swin\cite{dc-swin}         & 93.46    & 96.00    & 85.32   & 90.03 & 84.88 & 89.94         & 82.01       & 91.29    \\
					MPCNet\cite{wang2023mpcnet}         & 92.76    & 95.50    & 84.70   & 90.40 & 90.44 & 90.76         & 83.27       & 90.93    \\
					XNet\cite{xnet}         & 92.89    & 95.48    & 85.17   & 91.10 & 88.22 & 90.57         & 82.95       & 91.21    \\
					\midrule
					\textbf{SFFNet(Ours)}                  & \textbf{93.51}    & \textbf{96.25}    & \textbf{85.94}  & \textbf{91.43} & \textbf{91.24} & \textbf{91.67}         & \textbf{84.80}      & \textbf{91.91} 
					\\
					\bottomrule
					\bottomrule

			\end{tabular}}%
			\begin{tablenotes} 
				\item The best results are indicated in bold black. 
			\end{tablenotes} 
			\label{tab:5}%
		\end{threeparttable}
	\end{table*}%
	
	\begin{table*}[htbp]
		\centering
		\begin{threeparttable}
			\renewcommand\arraystretch{1.7}
			\centering
			\caption{Comparison of segmentation results on the Potsdam dataset.}
			\setlength{\tabcolsep}{4mm}{
				\begin{tabular}{l|ccccc|cccc}
					\toprule
					\toprule
					\multicolumn{1}{c|}{\multirow{2}{*}{\textbf{Method}}} & \multicolumn{5}{c|}{\textbf{F1(\%)}}                    & \multicolumn{3}{c}{\textbf{Evaluation   index}} \\\cline{2-5} \cline{6-9} 
					& \textbf{Imp.Surf} & \textbf{Building} & \textbf{Lowveg} & \textbf{Tree}  & \textbf{Car}   & \textbf{MeanF1(\%) }   & \textbf{mIoU(\%)}    & \textbf{OA(\%)}   \\
					\midrule
					ABCNet\cite{abcnet}              & 87.05    & 88.26    & 78.94  & 84.63 & 93.11 & 86.4          & 76.35       & 84.21    \\
					MAResU-Net\cite{resunet}         & 92.38    & 95.83    & 86.65  & 88.44 & 96.13 & 91.89         & 85.22       & 90.28    \\
					A2-FPN\cite{a2fpn}              & 92.85    & 95.84    & 87.17  & 88.76 & 96.13 & 92.15         & 85.66       & 90.68    \\
					ST-Unet\cite{stunet}         & 92.77    & 96.90    & 86.45   & 88.11 & 95.80 & 92.01         & 85.46       & 90.53    \\
					DC-Swin\cite{dc-swin}             & 93.26    & 96.86    & 87.74  & 88.68 & 95.50 & 92.41         & 86.10       & 91.16    \\
					MPCNet\cite{wang2023mpcnet}            & 92.69    & 96.38    & 87.30  &  88.74 &  96.34 &  92.29         & 85.91       & 90.56    \\
					XNet\cite{xnet}            & 91.98    & 96.80    & 86.93  &  88.76 &  95.81 &  92.05         & 85.51       & 90.56    \\
					\midrule
					\textbf{SFFNet(Ours)}            & \textbf{93.73}    &\textbf{97.02}    &\textbf{89.02}   & \textbf{90.26} & \textbf{96.79} &\textbf{93.36}         & \textbf{87.73}       &\textbf{91.88}   
					\\
					\bottomrule
					\bottomrule
			\end{tabular}}%
			\begin{tablenotes} 
				\item The best results are indicated in bold black. 
			\end{tablenotes} 
			\label{tab:6}%
		\end{threeparttable}
	\end{table*}%

	\subsection{Comparative Experiments}
	
	For the comparative models, the selected models are as follows: the "bilateral network" pure convolutional neural network ABCNet \cite{abcnet} with spatial and contextual pathways, multi-stage attention residual UNet (MAResU-Net) \cite{resunet} with linear attention mechanism, Attention Aggregation Feature Pyramid Network A2-FPN \cite{a2fpn}, Swin-Transformer network DC-Swin \cite{dc-swin} with Dense Connection Feature Aggregation Module (DCFAM), segmentation network ST-Unet \cite{stunet} with global-local feature fusion scheme, MPCNet \cite{wang2023mpcnet}, a network with multi-scale prototype transformer decoder. All of the above networks are spatial domain methods. We also compared with XNet \cite{xnet}, a network that directly utilizes frequency domain features for segmentation. Our model ultimately achieved higher accuracy on the ISPRS Vaihingen and ISPRS Potsdam datasets, which are widely used for remote sensing segmentation tasks, compared to the aforementioned models.
	
	\textbf{Results of the Vaihingen Dataset:} Tab.\ref{tab:5} presents numerical comparisons of various mainstream semantic segmentation methods on the Vaihingen dataset. The research findings indicate that our proposed SFFNet achieved performance of 91.67\% in average F1, 84.80\% in mIoU, and 91.91\% in OA, respectively. The best performance of SFFNet in F1, OA, and mIoU significantly outperformed other networks. SFFNet not only surpassed the excellent pure convolutional network ABCNet but also exceeded the DC-Swin network, based on Swintransformer, which has strong global information representation capability. Additionally, ST-Unet, currently a popular CNN combined with Transformer network, relies solely on spatial features. Our SFFNet outperformed ST-Unet by 3.41\% in mIoU, 2.07\% in F1, and 1.38\% in OA. XNet is a network that directly uses frequency domain features for segmentation. Our SFFNet surpassed XNet, which uses purely frequency domain features, by 1.1\% in mIoU, 1.85\% in F1, and 0.7\% in OA. This experiment demonstrates that our model has better segmentation performance.
	
	In addition to comparing the segmentation accuracy of each model, we also compared the convergence speed of mainstream models and our proposed SFFNet. The results, shown in Fig.\ref{fig:16}, depict the trend of mIoU changes with Epoch during the training of various models on the Vaihingen dataset. We selected the results of the first 44 rounds of training as reference, with the 15th epoch marking the end of the first cycle of cosine annealing strategy. From the results shown in Fig.\ref{fig:16}, our model stabilizes the fastest, indicating that SFFNet has better fitting ability compared to other networks. Furthermore, compared to other models, our model exhibits smaller fluctuations, suggesting that our model can more stably learn the features of tasks and datasets during the training process.
	
	Fig.\ref{fig:14} displays the segmentation results of SFFNet on ID 10 and 12 of the Vaihingen dataset. For targeted comparison, we provide locally enlarged images of the segmentation results of SFFNet and other advanced networks. The results are shown in Fig.\ref{fig:13}. From the first, second, and third rows of Fig.\ref{fig:13}, it can be seen that models such as ST-UNet and SFFNet with global modeling have better segmentation results for large-scale continuity and spatial correlation of features (e.g., buildings) compared to purely CNN-based networks such as ABCNet and MACU-Net. From the first, second, third, and fourth rows of the figure, it can be observed that methods with frequency domain features like XNet and SFFNet exhibit significantly better segmentation results in shadow areas (first row), edges (second and third rows), and regions with significant texture changes (fourth row) than pure spatial segmentation methods like ABCNet, MACU-Net, ST-UNet, and MPCNet. However, from the fifth row, it can be seen that XNet, which is solely based on frequency domain features, suffers from spatial information loss, as it splits the car in the ground into two parts. In contrast, our SFFNet not only retains sufficient spatial information but also incorporates additional frequency domain features. SFFNet performs better than pure spatial segmentation methods in areas with large grayscale variations while maintaining sufficient spatial information.

	\textbf{Results on the Potsdam Dataset:} To comprehensively evaluate the network performance, we conducted further experiments on the Potsdam dataset. As shown in Tab.\ref{tab:6}, SFFNet achieved remarkable results on the Potsdam test set: with an average F1 score of 93.36\%, mIoU of 87.73\%, and OA index of 91.88\%, all surpassing other methods. Due to differences in dataset size and type, segmentation accuracy on the Potsdam dataset is generally higher than that on the Vaihingen dataset. Compared to pure spatial segmentation methods, SFFNet yielded superior results. Compared to ST-Unet, which incorporates both spatial global and local features, our SFFNet showed better performance after integrating frequency domain features, with F1, mIoU, and OA scores respectively surpassing ST-Unet by 1.35\%, 2.27\%, and 1.35\% on this dataset. Additionally, our method outperformed XNet, a method solely using frequency domain features, by 1.31\%, 2.22\%, and 1.32\% in terms of F1, mIoU, and OA on this dataset.
	
	As illustrated in Fig.\ref{fig:17}, we also provide overall segmentation images for IDs 3\_13 and 3\_14. Fig.\ref{fig:18} displays the segmentation results of the network models mentioned in Tab.\ref{tab:6} on the Potsdam dataset. From the second row of Fig.\ref{fig:18} and the first and third rows, it can be observed that networks with global modeling capabilities, such as SFFNet, DC-Swin, and ST-Unet, outperform networks without global dependencies, such as ABCNet and MACU-Net, especially in segmenting large continuous areas such as buildings and streets. From the first, second, third, and fourth rows of Fig.\ref{fig:18}, networks like XNet and our SFFNet, which utilize frequency domain information, perform better in segmenting areas with significant grayscale variations compared to all pure spatial segmentation networks, such as shadows (first row), edges (second and third rows), and high-texture regions (fourth row). However, as shown in the fifth row of Fig.\ref{fig:18}, XNet suffers from spatial information loss, as seen in mislabeling trees as buildings. In contrast, our SFFNet demonstrates superior spatial awareness while leveraging the advantages of frequency domain features to address the shortcomings of pure spatial segmentation, thereby improving the accuracy and robustness of remote sensing image segmentation.
	
	By conducting experiments on both the Vaihingen and Potsdam datasets, we have demonstrated the excellent performance of SFFNet on both datasets, showing significant segmentation capabilities. SFFNet preserves sufficient spatial information in features while leveraging frequency domain features to improve the shortcomings of pure spatial methods.

	

	\section{Conclusion}
	
	This paper proposes a two-stage segmentation network called SFFNet, which addresses the challenge of accurately segmenting areas with large grayscale variations while preserving sufficient spatial information by fully considering both frequency domain and spatial domain features. Specifically, to retain the spatial information of spatial domain features, the first stage involves the extraction of these features. Then, in the second stage, spatial domain feature mapping (including global and local spatial features) is performed to preserve more extensive spatial information. Furthermore, to achieve more efficient global mapping in remote sensing applications, Swintransformer is enhanced in this study to better adapt to remote sensing tasks. To obtain additional frequency domain information through frequency domain feature mapping (including low-frequency and high-frequency features), a WTFD (Wavelet Transform Feature Decomposer) is designed, utilizing Haar wavelet transform for frequency domain feature mapping. Subsequently, efforts are made to facilitate the integration of low-frequency features with global features and high-frequency features with local features, bridge the semantic gap between frequency domain and spatial domain features, and perform feature selection, achieved through the design of a MDAF (Multiscale Dual-Representation Alignment Filter). This structure employs multiscale vertical convolution for feature multiscale expansion, followed by the utilization of a DAF (Dual-Representation Alignment Filter) based on cross-attention for semantic alignment and ultimately feature selection. Experimental results demonstrate the superiority of the SFFNet structure and the effectiveness of each component. The paper aims to inspire more researchers to propose practical solutions to address the challenge of segmenting areas with large grayscale variations and encourages further exploration of the potential and applications of Haar wavelet transform decomposer in the field of remote sensing.

	\bibliography{ref}   
	\bibliographystyle{spiebib}   
	\ifCLASSOPTIONcaptionsoff
	\newpage
	\fi

\end{document}